\def\year{2020}\relax

\documentclass[letterpaper]{article} 
\usepackage{aaai20}  
\usepackage{times} 
\usepackage{helvet}
\usepackage{courier}  
\usepackage[hyphens]{url} 
\usepackage{graphicx} 
\usepackage{graphicx}  
\nocopyright

\usepackage{epsfig}
\usepackage{amsmath}
\usepackage{amssymb}

\usepackage{booktabs}
\usepackage{graphicx}
\usepackage{algorithm}
\usepackage{algorithmic}
\usepackage{adjustbox}
\usepackage{etex}
\usepackage{bm}
\usepackage{bbm}
\usepackage[mathscr]{eucal}
\usepackage{epstopdf}
\usepackage{makecell}
\usepackage{multirow}
\usepackage{wrapfig}
\usepackage{subcaption}
\usepackage{javen}
\usepackage{threeparttable}
\usepackage{tablefootnote}
\usepackage{soul}
\usepackage{comment}

\DeclareMathAlphabet{\mathcal}{OMS}{cmsy}{m}{n}

\title{Multi-objective Neural Architecture Search\\
via Non-stationary Policy Gradient}

\author{Zewei Chen, Fengwei Zhou, George Trimponias, Zhenguo Li\\ 
Huawei Noah's Ark Lab\\
\{chen.zewei, zhoufengwei, g.trimponias, li.zhenguo\}@huawei.com
}
 
\begin{document}

\maketitle

\begin{abstract}

Multi-objective Neural Architecture Search (NAS) aims to discover novel architectures in the presence of multiple conflicting objectives. 
Despite recent progress, the problem of approximating the full Pareto front accurately and efficiently remains challenging. 
In this work, we explore the novel reinforcement learning (RL) based paradigm of non-stationary policy gradient (NPG). NPG utilizes a non-stationary reward function, and encourages a continuous adaptation of the policy to capture the entire Pareto front efficiently. We introduce two novel reward functions with elements from the dominant paradigms of scalarization and evolution.
To handle non-stationarity, we propose a new exploration scheme using cosine temperature decay with warm restarts. 
For fast and accurate architecture evaluation, we introduce a novel pre-trained shared model that we continuously fine-tune throughout training. 
Our extensive experimental study with various datasets shows  that  our  framework can approximate the full Pareto front well at fast speeds. Moreover, our discovered cells can achieve supreme predictive performance compared to other multi-objective NAS methods, and other single-objective NAS methods at similar network sizes. Our work demonstrates the potential of NPG as a simple, efficient, and effective paradigm for multi-objective NAS.

\end{abstract}

\section{Introduction}
Neural Architecture Search (NAS) automatically designs neural architectures, which is otherwise a time-consuming and labor-intensive process~\cite{zoph2017neural,baker2017designing,zoph2018learning,pham2018efficient,brock2018smash,zhang2018graph,liu2018darts,xie2018snas,cai2018proxylessnas}.
Traditionally, NAS searches for architectures with maximal predictive performance. 
However, in real applications, additional objectives such as inference time and energy consumption must be considered.
The trade-off among different objectives is typically captured by the Pareto front, i.e., the set of Pareto-optimal architectures with the property that no objective can be improved without harming the other objectives.

\begin{figure}[t]
    \centering
    \includegraphics[scale=0.33]{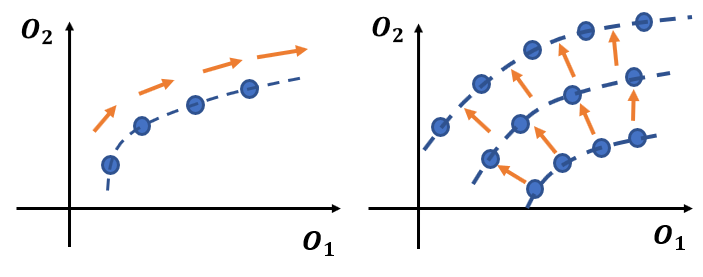}
    \caption{Multi-objective NAS for maximizing accuracy ($O_2$) and minimizing number of parameters ($O_1$). 
    ADF expands the Pareto front \textbf{(left)} from left to right.
    ADC expands it in bands \textbf{(right)}.
    }
    \label{fig:algorithm_illustrate}
\end{figure}

Obtaining the full Pareto front accurately and efficiently is challenging for NAS, because the number of possible architectures is typically very large and architecture evaluation is expensive. Most works on multi-objective NAS approximate the Pareto front by relying on the two paradigms of scalarization and evolution. Scalarization combines the multiple objectives into a single one (e.g., by using a weighted sum or product). 
However, multiple runs with different scalarizations are required to obtain multiple Pareto-optimal solutions, which is computationally-intensive. 
Evolution involves the use of genetic algorithms
~\cite{srinivas1994muiltiobjective,deb2000fast}, 
which iteratively update the current population to create a new generation via crossovers, mutations and elitist selection. A recent genetic approach is LEMONADE~\cite{elsken2018efficient}, which can approximate the Pareto front well but at a relatively high search cost. NSGA-Net~\cite{lu2018nsga}, a different evolutionary method, achieves comparatively lower cost by training sampled architectures for only 25 epochs, which is nevertheless a low fidelity estimate with high bias~\cite{survey}.
This, in turn, can take a toll on the ability to approximate the true Pareto front well.

A critical observation on the two aforementioned para-digms is that it is in principle possible to continuously adapt a single policy to efficiently generate Pareto-optimal architectures. For scalarization, we can adapt the optimal policy for a given scalarization to obtain an optimal policy for a similar scalarization. For evolution, we can
continuously change the policy to produce architectures that Pareto-dominate existing ones.
Inspired by this observation, our work explores the novel paradigm of non-stationary policy gradient (NPG) for approximating efficiently and accurately the full Pareto front.
NPG is built upon an RL framework, and is compatible with existing RL-based NAS~\cite{pham2018efficient}. 
We opt for RL, since it is well-established in traditional NAS and because the reward is easy to customize.
In detail, a controller learns a policy to generate architectures. Unlike traditional RL though, the reward function is non-stationary;
this allows us to efficiently produce the entire Pareto front by continuously adapting the policy according to the changing reward.

NPG-NAS, our proposed framework, introduces two new reward functions.
The first one (ADF) utilizes a target-based desirability function \cite{derringer1980simultaneous} to scalarize the objectives. 
By slowly annealing target values for all but one objectives, it gradually uncovers the entire Pareto front in a single run. 
The second one (ADC) is based on the Pareto dominance concept: an architecture receives a reward based on whether it dominates or is dominated by other architectures in the current Pareto front. 
As shown in Fig. \ref{fig:algorithm_illustrate}, ADF traverses the entire Pareto front, whereas ADC expands it in bands like evolution. 
ADF relies on scalarization, while ADC is based on Pareto dominance which is ubiquitous in evolutionary methods.
But while evolution typically creates a new generation via genetic operations on the best-fit individuals of the current generation as measured by a fitness function, NPG-NAS expands the band by sampling new individuals that are likely to dominate solutions in the current Pareto front using a non-stationary policy that learns from observed non-stationary rewards.

To deal with non-stationarity, we propose a new exploration scheme with a changing Boltzmann temperature to better balance exploration and exploitation, which is inspired by prior work on multi-armed bandits~\cite{Besbes:2014}. 
Furthermore, a common challenge for NAS concerns architecture evaluation.
Techniques such as parameter sharing~\cite{pham2018efficient}, continuous relaxation~\cite{liu2018darts}, and network morphisms~\cite{Wei:2016:NM:3045390.3045451} have addressed this. 
For fast and accurate evaluation, our work introduces a pre-trained shared model which is continuously fine-tuned during search whenever an architecture is sampled. Our pre-trained shared model leads to fast convergence in just 50 steps, as well as high correlation between the validation accuracy during search versus when architectures are trained from scratch.

Our extensive experimental study on CIFAR-10 and CIFAR-100 based on a cell-based search space and with a strong high-budget baseline shows that NPG-NAS can approximate the full Pareto front accurately at fast speeds. Furthermore, by stacking our discovered cells we obtain networks with very high predictive performance. 
For LEMONADE and NSGA-Net, the performance from transferring cells searched on CIFAR-10 to other datasets (ImageNet or CIFAR-100) is less striking than their performance on CIFAR-10.
For NPG-NAS, we transfer architecture cells found for CIFAR-10 and CIFAR-100 to ImageNet, yielding higher classification accuracies than most multi-objective methods, or competitive accuracies at a fraction of the GPU cost. 
Importantly, both our searched and transferred cells are often on par with or even outperform many state-of-the-art NAS methods with similar network sizes. 
This points to the high quality of our discovered cells. 
Our experiments unveil fundamental differences in the way architectures are sampled for ADF and ADC versus random search. We show that random search performs poorly in the multi-objective setting, contrary to single-objective NAS~\cite{yang2019nas}.

NPG-NAS is a fully gradient-based scheme, as it uses gradient descent to optimize both the network weights and the controller policy. It favors a simple design, albeit based on intuitive and sound fundamentals. Furthermore, it is a lightweight framework, because it only maintains a single policy that is continuously updated throughout training. 
Its high efficiency and effectiveness across various datasets make it a well-suited framework for multi-objective NAS.
\section{Related Work}

\subsection{Multi-objective NAS}
A comprehensive survey on single- and multi-objective NAS is given in~\cite{survey}.
Most works on multi-objective NAS can be divided in two classes\footnote{We only survey multi-objective NAS here. Interested readers can refer to \cite{Marler2004,Roijers:2013} for general surveys on multi-objective optimization or multi-objective MDPs.}. Methods in the first-class scalarize the multiple objectives into a single one by using their weighted sum or product, and subsequently solve a single-objective optimization problem.
For non-differentiable objectives, a continuous relaxation technique enables gradient-based optimization, e.g., FBNet~\cite{wu2018fbnet}. For RL-based MnasNet~\cite{tan2018mnasnet} and MONAS~\cite{hsu2018monas}, 
the reward is a scalarized objective, where the trade-off among various objectives is controlled by tuning the corresponding weights.
The second class includes genetic algorithms with non-dominated sorting.
These methods sort the population into a hierarchy of sub-populations based on the ordering of Pareto dominance.
Elitist selection criteria exploit this ordering to guide the evolution process and discover diverse Pareto-optimal architectures \cite{srinivas1994muiltiobjective,deb2000fast}.
LEMONADE~\cite{elsken2018efficient} and NSGA-Net~\cite{lu2018nsga} fall into this category. 
Note that a few multi-objective NAS works are based on alternative paradigms; DPP-NET~\cite{dong2018dpp} for instance is based on sequential model-based optimization~\cite{10.1007/978-3-642-25566-3_40}. In this work, we focus on the dominant paradigms of scalarization and evolution that our work draws inspiration from.

\subsection{Policy Gradient for Multi-objective MDPs}
Policy gradient methods for multi-objective RL have received some attention in the prior literature.
Two relevant approaches are (i) the radial and (ii) the Pareto-following \cite{6889738}. 
Our first method ADF shares similarities with the Pareto-following approach: they both start with a Pareto-optimal policy, and subsequently move along the Pareto front. But instead of using update and correction steps, ADF relies on a simpler policy transfer scheme, whereby the policy of a current target is transferred and adapted to the next target. 
A different gradient-based approach in \cite{Parisi:2016} learns a direction to update the policy manifold in order to improve the Pareto front approximation. Rather than trying to learn the entire policy manifold at once, NPG continuously adapts a single policy to efficiently generate the full Pareto front.

\subsection{Non-stationary Rewards}
Non-stationarity has attracted interest in the multi-armed bandit (MAB) and RL literature.
Prior RL works typically deal with general non-stationary environments, e.g., \cite{Goldberg2003,daSilva:2006,5137416,Al-ShedivatBBSM18}.
A relevant approach for efficient policy adaptation is to meta-learn the changes in the reward and update the policy accordingly~\cite{Al-ShedivatBBSM18}.
In our work, we opt for a temperature-based exploration scheme because it is very simple yet effective.
It is loosely inspired by prior work on the stochastic MAB setting with non-stationary rewards, which highlights the inherent trade-off between remembering and forgetting \cite{Besbes:2014}. Old information may become obsolete, and negatively affect adaptation to the changing environment. For this reason, the authors propose a mechanism that forgets any acquired information at regular intervals, and restarts anew.
Given our goal of continuous policy adaptation, our scheme does not
erase the knowledge present in the policy. Instead, we opt for a softer mechanism that increases exploration at regular intervals while exploiting what is currently encoded in the policy.
To our knowledge, non-stationary rewards have not been previously explored for NAS.
Interestingly, our novel exploration scheme may be relevant in other settings involving non-stationary rewards, even outside NAS.
\section{Background}
Without loss of generality about $\min$ or $\max$, given $m\geq2$ objective functions $f_1: \mathcal{X}\rightarrow\mathbb{R},\dots,f_m: \mathcal{X}\rightarrow\mathbb{R}$, a multi-objective optimization problem has the form:
\begin{equation}
\label{eq:MOP}
\max f_1(\bm{x}),\dots,\max f_m(\bm{x})\nonumber\text{ s.t. } \bm{x}\in\mathcal{X}.
\end{equation}
$\mathcal{X}$ is the feasible set of decision vectors.
As there may be no solution that simultaneously optimizes all $m$ objectives, the primary solution concept is \textit{Pareto optimality}:  
$\bm{x}_1\in\mathcal{X}$ dominates $\bm{x}_2\in\mathcal{X}$, if (i) $f_i(\bm{x}_1)\geq f_i(\bm{x}_2), \forall i\in\{1,\dots,m\}$, and (ii) $f_j(\bm{x}_1)> f_j(\bm{x}_2)$ for at least one $j\in\{1,\dots,m\}$. In that case, we write $\bm{x}_1\succ\bm{x}_2$.
A decision vector $\bm{x}^*\in\mathcal{X}$ is called Pareto optimal, if there is no $\bm{x}\in\mathcal{X}$ that dominates $\bm{x}^*$.
The Pareto front $\mathcal{P}_f$ is the set of all Pareto-optimal solutions.

\noindent\textbf{Desirability Function.}
The desirability function is used in the optimization of multiple objective engineering systems to set desired limits for the decision vector \cite{derringer1980simultaneous}.
Our work focuses on the ``target is best'' type, where the goal is for the objective to be near a given target.

\noindent\textbf{RL for Single-Objective NAS.}
Prior work on RL-based NAS \cite{zoph2017neural,pham2018efficient} considers a recurrent neural network (RNN) controller, parameterized by $\bm{\theta}\in\mathbb{R}^d$, which samples child 
architectures $\bm{\alpha}$ based on a policy $\pi(\bm{\alpha};\bm{\theta})$. The goal is to compute the optimal policy $\pi^*=\pi(\bm{\alpha};\bm{\theta}^*)$. Since conventional NAS is interested in architectures with high accuracy, the reward $\mathcal{R}(\bm{\alpha})$ is equal to the validation accuracy ${\rm{acc_{valid}}}(\bm{\alpha})$ of child model $\bm{\alpha}$. Given that the reward signal is non-differentiable, policy gradient methods are used to iteratively update $\bm{\theta}$. In practice, a mini-batch of several architectures is sampled according to $\pi(\bm{\alpha};\bm{\theta})$ in order to approximate the quantity $\mathbb{E}_{\bm{\alpha}\sim\pi(\bm{\alpha};\bm{\theta})}[\mathcal{R}(\bm{\alpha})]$. 
We use the REINFORCE method with baseline \cite{Williams:1992:SSG:139611.139614}, because we empirically find it performs well, but any policy gradient method may be used.

\section{Algorithms}
\label{sec:NPG-NAS-Alg}

\subsection{Annealing  Desirability Function (ADF)}
\label{sec:ADF}
Our first reward function for NPG-NAS utilizes a desirability function. We start by discussing the two-objective case. The first objective $f_1$ is the validation accuracy, while the second objective $f_2$ can be any other objective of interest.

\noindent\textbf{Two objectives.}
Assume a desirability function of the ``target is best'' type for $f_2$ of the following \textit{triangular} form:
\begin{equation}
\label{eq:triang}
    \mathcal{F}_2(\bm{\alpha};\tau,\delta)=
    \begin{cases}
        1-\frac{|\tau-f_2(\bm{\alpha})|}{\delta}, \text{ if } |\tau-f_2(\bm{\alpha})|\leq\delta\\
        0, \text{ otherwise}.
    \end{cases}
\end{equation}
Eq. \eqref{eq:triang} defines a piecewise linear function with a maximum value of 1 at $f_2(\bm{\alpha})=\tau$, which decreases linearly at a rate of $\frac{1}{\delta}$ until it becomes 0 when $f_2(\bm{\alpha})=\tau-\delta$ or $f_2(\bm{\alpha})=\tau+\delta$. 
Based on \eqref{eq:triang}, we define the multiplicative reward function:
\begin{equation}
\label{eq:DF-reward}
    \mathcal{R}(\bm{\alpha};\tau,\delta) = {\rm{acc_{valid}}}(\bm{\alpha})\cdot \mathcal{F}_2(\bm{\alpha};\tau,\delta).
\end{equation}
Eq. \eqref{eq:DF-reward} assigns higher rewards to architectures $\bm{\alpha}$ (i) with high validation accuracy, and (ii) whose second objective $f_2(\bm{\alpha})$ is close enough to target $\tau$. 
Ideally, we want to discover the best architecture for each possible target for $f_2$.
How to do this?

For simplicity, let $\mathcal{T}=[\tau_{min},\tau_{max}]$, i.e., the set of targets takes values in a continuous range between a minimum target $\tau_{min}$ and a maximum target $\tau_{max}$. 
Assume we know the optimal controller policy for target $\tau$. 
To compute the optimal policy $\pi'$ for a new target $\tau'$ close to $\tau$, we can start from the known optimal policy $\pi$ for $\tau$, and just do a few update steps to get the optimal policy for $\tau'$. Assuming the optimal policy changes relatively smoothly in the neighborhood of $\tau$, a few samples should suffice to learn the optimal policy $\pi'$ for $\tau'$. 
If $\theta^*$ is the optimal policy parameter at target $\tau$, we can hence claim that the optimal policy parameter $\theta'^*$ at a near target $\tau'$ will be near $\theta^*$.

\begin{algorithm}[t!]\small
    \caption{ADF (\underline{A}nnealing \underline{D}esirability \underline{F}unction)}
    \label{algo:DF}
    \begin{algorithmic}[1]
        \REQUIRE datasets $\mathcal{D}_{train}$ and $\mathcal{D}_{valid}$, minimum target $\tau_{min}$, maximum target $\tau_{max}$,
        annealing steps $N_{anneal}$, warm-up steps $N_{warm}$, warm-up width $\delta_{warm}$, annealing width $\delta_{annealing}$.
        \ENSURE Pareto front $\mathcal{P}_f$.
        \STATE Randomly initialize RNN controller weight $\bm{\theta}$;
        \STATE Initialize set of sampled architectures $\mathcal{A}\leftarrow\emptyset$;
        \STATE \textit{//Warm-up Phase}
        \STATE Fix target $\tau\leftarrow\tau_{min}$;
        \STATE Fix width $\delta\leftarrow\delta_{warm}$;
        \FOR {$step=1$ \textbf{to} $N_{warm}$}
        \STATE Sample a child model $\bm{\alpha}_{ch}\sim\pi(\bm{\alpha};\bm{\theta})$;
        \STATE Evaluate $\bm{\alpha}_{ch}$ and get validation accuracy ${\rm{acc_{valid}}}(\bm{\alpha}_{ch})$;
        \STATE Compute reward according to Eq. \eqref{eq:DF-reward};
        \STATE Update $\bm{\theta}$ using REINFORCE;
        \ENDFOR
        \STATE \textit{//Annealing Phase}
        \STATE Fix width $\delta\leftarrow\delta_{anneal}$;
        \FOR {$step=1$ \textbf{to} $N_{anneal}$}
        \STATE Set target $\tau\leftarrow\tau_{min}+\frac{\tau_{max}-\tau_{min}}{N_{anneal}}\cdot step$;
        \STATE Perform Lines 7-10;
        \STATE $\mathcal{A}\leftarrow\mathcal{A}\cup\{(\bm{\alpha}_{ch},({\rm{acc_{valid}}}(\bm{\alpha}_{ch}),f_2(\alpha_{ch})))\}$;
        \ENDFOR
        \STATE Extract Pareto front $\mathcal{P}_f$ from $\mathcal{A}$;
    \end{algorithmic}
\end{algorithm}

We show ADF in Algorithm \ref{algo:DF} which consists of two phases. Both phases sample a child architecture $\bm{\alpha}_{ch}$ in each step according to policy $\pi(\bm{\alpha};\bm{\theta})$.
Using our shared model (see Section \ref{sec:pretrained-model}), we determine ${\rm{acc_{valid}}}(\bm{\alpha}_{ch})$ and compute reward \eqref{eq:DF-reward}. We use REINFORCE to update $\bm{\theta}$ (Lines 7-10 and Line 16). 
The two phases serve different purposes. 
The first one is a warm-up phase with $N_{warm}$ steps (Lines 3-11). It learns an initial optimal policy for the leftmost target $\tau_{min}$ to start the subsequent annealing process. 
It uses a fixed width $\delta_{warm}$ (Line 5) of relatively high value; we use  $\delta_{warm}=\frac{\tau_{max}-\tau_{min}}{2}$. 
The second phase is the annealing phase that slowly increases the target $\tau$ from $\tau_{min}$ to $\tau_{max}$ at a linear rate over a total of $N_{anneal}$ steps (Lines 12-18). In this phase, the width is set to a lower value $\delta_{anneal}$ (Line 13); we choose $\delta_{anneal}=\frac{\tau_{max}-\tau_{min}}{10}$. This phase gradually learns the optimal policy for all annealed targets, so we use a smaller width to limit the search process near the target.
At the end of the annealing phase we extract the Pareto front from the sampled set $\mathcal{A}$ (Line 19).

\noindent\textbf{Generalizing to more objectives.}
For $m\geq3$ we simultaneously anneal the targets for $m-1$ of the objectives, say $f_2,\dots,f_m$. To this goal, we first create a grid containing target pairs for $(f_2,\dots,f_m)$, and define a $(m-1)$-D desirability function for each target. Subsequently, we traverse this grid using any space-filling curve that can preserve locality, so that any two consecutive points that the curve passes through are close in the $(m-1)$-D grid.

\subsection{Assigning Dominance-based Credit (ADC)}
\label{sec:NPG-NAS-NDS}
In this section, we propose our second reward function. 
We first define some relevant concepts. Given the current Pareto front $\mathcal{P}_f$ and an architecture $\bm{\alpha}$, we define (i) $N(\mathcal{P}_f)$ as the number of points in $\mathcal{P}_f$, (ii) $N^{\succ}(\bm{\alpha};\mathcal{P}_f)$ as the number of architectures in $\mathcal{P}_f$ that dominate $\bm{\alpha}$, and $N^{\prec}(\bm{\alpha};\mathcal{P}_f)$ as the number of architectures in $\mathcal{P}_f$ that are dominated by $\bm{\alpha}$.
Furthermore, given a radius $\epsilon$ we define the density function $\rho(\bm{\alpha};\mathcal{P}_f,\epsilon)$ as the number of points in $\mathcal{P}_f$ that lie within distance $\epsilon$ from $\mathbf{f}(\bm{\alpha})=(f_1(\bm{\alpha}),\dots,f_m(\bm{\alpha}))$. The distance can be defined based on $L_1$ or $L_2$ norms.
The reward  is:
\begin{align}
\label{eq:reward}
  &\mathcal{R}(\bm{\alpha}) = \mathcal{R}(\bm{\alpha};\mathcal{P}_f, \epsilon, C)=\\
  &\begin{cases}
    -\tanh\left(\frac{N^{\succ}(\bm{\alpha};\mathcal{P}_f) + \rho(\bm{\alpha};\mathcal{P}_f,\epsilon)}{C}\right), \text{ if }\exists\bm{\alpha}_{p} \in \mathcal{P}_f: \bm{\alpha}_{p} \succ \bm{\alpha}, \\
    \tanh\left(\frac{N(\mathcal{P}_f)+N^{\prec}(\bm{\alpha};\mathcal{P}_f)}{C}\right), \text{ otherwise}.\nonumber
  \end{cases}
\end{align}
\noindent Eq. \eqref{eq:reward} is non-symmetric about the current front $\mathcal{P}_f$. 
In \eqref{eq:reward}, the more points in the Pareto front a newly sampled architecture $\bm{\alpha}$ dominates, the higher its assigned reward.
This is due to the added difficulty of discovering new Pareto points that dominate (several) existing points on $\mathcal{P}_f$. 
Furthermore, the reward is higher as the number $N(\mathcal{P}_f)$ of points in the Pareto front increases, because it gets more difficult to discover new Pareto-optimal points.
If $\bm{\alpha}$ is dominated by $\mathcal{P}_f$, it receives a negative reward based on (i) the number of points in $\mathcal{P}_f$ that dominate it, and (ii) its density.
A larger density term $\rho(\bm{\alpha};\mathcal{P}_f,\epsilon)$ incurs a penalty to discourage re-sampling from oversampled subspaces. 
$C>0$ is a tunable hyperparameter to scale the numerator so that the $\tanh$ operator is saturated at a desired threshold.

\begin{algorithm}[t!]\small
    \caption{ADC (\underline{A}ssigning \underline{D}ominance-based \underline{C}redit)}
    \label{algo:NDS}
    \begin{algorithmic}[1]
        \REQUIRE datasets $\mathcal{D}_{train}$ and $\mathcal{D}_{valid}$, radius $\epsilon>0$, controller steps $N_{steps}$, hyperparameter $C>0$.
        \ENSURE Pareto-optimal front $\mathcal{P}_f$.
        \STATE Randomly initialize RNN controller weight $\bm{\theta}$;
        \STATE $\mathcal{P}_f\leftarrow\emptyset$;
        \FOR {$step=1$ \textbf{to} $N_{steps}$}
        \STATE Sample a child model $\bm{\alpha}_{ch}\sim\pi(\bm{\alpha};\bm{\theta})$;
        \STATE Evaluate child model $\bm{\alpha}_{ch}$ and compute its $m$ objectives;%
        \STATE {Compute reward according to Eq. \eqref{eq:reward}};
        \STATE $dominated\leftarrow FALSE$;
        \FOR{each $\bm{\alpha}_{p} \in \mathcal{P}_f$}
        \IF {$\bm{\alpha}_{ch} \succ \bm{\alpha}_{p}$}
        \STATE $\mathcal{P}_f\leftarrow\mathcal{P}_f-\bm{\alpha}_{p}$;
        \ELSIF {$\bm{\alpha}_{p} \succ \bm{\alpha}_{ch}$}
        \STATE $dominated\leftarrow TRUE$;
        \ENDIF
        \ENDFOR
        \IF{$dominated == FALSE$}
            \STATE $\mathcal{P}_f\leftarrow\mathcal{P}_f\cup\{\bm{\alpha}_{ch}\}$;
        \ENDIF
        \STATE Update $\bm{\theta}$ using REINFORCE;
        \ENDFOR
    \end{algorithmic}
\end{algorithm}

We describe ADC in Algorithm \ref{algo:NDS}.
Similar to ADF, the controller samples in each step an architecture $\bm{\alpha}_{ch}$ based on $\pi(\bm{\alpha};\bm{\theta})$ (Line 4), for a total of $N_{steps}$ steps. 
The child model $\bm{\alpha}_{ch}$ is trained and subsequently, its $m$ objectives are computed (Line 5).
Next, we compute the reward based on Eq. \eqref{eq:reward} (Line 6).
If $\bm{\alpha}_{ch}$ dominates any $\bm{\alpha}_{p} \in \mathcal{P}_f$, then it is added into $\mathcal{P}_f$ while $\bm{\alpha}_{p}$ is removed (Lines 7-17).
If $\bm{\alpha}_{ch}$ is dominated by at least one $\bm{\alpha}_{p} \in \mathcal{P}_f$, 
$\mathcal{P}_f$ remains unchanged. 
If $\bm{\alpha}_{ch}$ is neither dominated by nor dominates any $\bm{\alpha}_{p} \in \mathcal{P}_f$, then it is added into $\mathcal{P}_f$.
Finally, we update the policy network $\bm{\theta}$ using REINFORCE (Line 18).

\section{Optimizations}
\subsection{Addressing Non-stationarity}
\label{sec:non-stationary}
NPG-NAS uses non-stationary rewards. ADF gradually anneals the target, changing reward function definition \eqref{eq:DF-reward}.
ADC continuously updates the current Pareto front $\mathcal{P}_f$, which renders reward function \eqref{eq:reward} non-stationary. 
The non-stationary nature of the reward function is hardly surprising. In multi-objective RL, the Pareto front does not consist of a single but of multiple policies \cite{Roijers:2013}. 
A single stationary reward would be insufficient to produce the diverse set of policies in the Pareto front.
Non-stationarity allows us to approximate the full Pareto front by a single policy that we continuously adapt. We thus do not need to simultaneously learn and maintain several policies.

Traditional RL manages the exploration/exploitation trade-off, by exploring more in the early stages of learning and exploiting more in the later stages.
Non-stationarity complicates this, since the changing rewards may make older knowledge obsolete, and the agent needs to continue to learn by interacting with the environment.
We address the challenge by exploiting a temperature parameter $T$, as in Boltzmann exploration~\cite{sutton2011reinforcement}. 
We propose to use cosine temperature decay with warm restarts\footnote{A similar scheme for learning rate schedules was introduced in \cite{loshchilov-ICLR17SGDR}.}. Our scheme starts with a high temperature $T_{max}$, which is slowly decreased within $\nu+1$ steps to a low value $T_{min}$ using cosine decay. Upon reaching $T_{min}$, we reset the temperature to $T_{max}$ and repeat the process.
Increasing the temperature introduces randomness in the policy and helps the controller explore different actions. Decreasing $T$ allows the controller to exploit its observations and settle down to a new policy.
The discontinuous jump (warm restart) helps to attenuate the impact of the old policy faster. 
We write:
\begin{equation*}
 T=T_{min}+\frac{T_{max}-T_{min}}{2}\cdot\bigg(1+\cos\big(\pi\frac{step\text{ mod }(\nu+1)}{\nu}\big)\bigg).
\end{equation*}

Our scheme is inspired by the Rexp3 algorithm in stochastic multi-armed bandits with non-stationary rewards \cite{Besbes:2014}, which discards previously acquired information at regular intervals to ensure that the policy is not negatively impacted by the changing rewards. 
However, instead of forgetting everything at regular intervals, we adopt a softer approach whereby we increase exploration while also being able to exploit the knowledge currently present in the policy. 
This is also consistent with our policy adaptation goal.

\subsection{Pre-trained Shared Model}
\label{sec:pretrained-model}
ENAS~\cite{pham2018efficient} introduced parameter sharing for fast architecture evaluation. However, its child models exhibit different performance during the early and late stages of the search process, while NPG-NAS requires the ability to accurately evaluate a child model at any point of search. We thus propose a novel pre-trained shared model. 
Our model uses the super network of ENAS, where the weights for each operation are shared by all child models with this operation. But unlike ENAS, we introduce a pre-training phase and a different update mechanism, which allows us to evaluate child models accurately from the early stages of search.

Pre-training a shared super network is different from pre-training a single model. The pre-training procedure must satisfy the following requirements: 1) each operation should be treated equally, and 2) the weights of any operation should not favor particular models. In this spirit, 
we pre-train the super network as follows: (i) Initialize shared weights randomly; (ii) Sample a child model randomly; (iii) Sample a batch of training images and update parameters of the sampled model by one-step gradient descent; (iv) Repeat (ii) and (iii) until the performance on a batch of 64 randomly sampled architectures converges.
By randomly sampling child models we can sample fairly all operators of the computational graph; and by using one gradient step we avoid overfitting the pre-trained shared parameters to any specific architecture.
During search, we initialize the shared super network with the pre-trained parameters. 
To evaluate a sampled child model, we fine-tune the shared weights using just a few gradient steps.
This is different from ENAS, which evaluates a child model directly on the shared model, without fine-tuning it first. 
As shown in Section \ref{sec:exp-results}, our pre-trained shared model has two desirable properties: (1) a few gradient steps, e.g., 30 to 50, suffice for convergence of the weights of the child model, and (2) the fine-tuned accuracies are highly correlated with the stand-alone ones. 

Finally, note that it would also be possible to use meta-learning to learn good weights (initialization) for the shared model~\cite{10.5555/3305381.3305498,li2017metasgd}, in order to be able to converge to the correct shared weights for the sampled architectures using few gradient steps only. 
In our work, we instead opt for a pre-trained model that we continuously fine-tune.
We adopt this simpler scheme, which does not require a separate meta-training phase but produces strong results, as we discuss in Section \ref{sec:exp-results}.
Interestingly, our pre-trained model has a similar effect as meta-learning algorithms since it leads to fast convergence for any sampled architecture using a few gradient steps only (e.g., 30 to 50).
\section{Experimental Studies}
\label{sec:exp-studies}

\subsection{Implementation}
\label{sec:exp-design}
\textbf{Datasets.} We search architecture cells on CIFAR-10 and CIFAR-100~\cite{krizhevsky2009learning}, which we then transfer to ImageNet. For the search phase, the original training set is divided into a set $\mathcal{D}_{train}$ with 45000 images to train the sampled architectures, and a set $\mathcal{D}_{valid}$ with 5000 images to evaluate them. To evaluate the finally selected architectures, we use the original training set to train the architectures from scratch and evaluate them on the test set. 

\noindent\textbf{Objectives.} We experiment with two and three objectives: maximizing the validation accuracy, minimizing the number of parameters, and minimizing number of FLOPs (for 3 objectives only). Hardware-aware objectives such as inference time and energy consumption are important for real applications; we choose nevertheless hardware-agnostic objectives without loss of generality for simplicity and reproducibility.

\noindent\textbf{Search space.} 
We opt for the cell-based search space of ENAS because it is flexible, easily transferable, and very effective (e.g., ENAS, DARTS).
Concretely, we first search for normal and reduction cells, and then stack them to construct a network. Note that this search space does not include searching over the width or depth of the neural net. Despite this limitation, cell-based approaches can yield competitive results \cite{pham2018efficient}.
We manually find that the minimum and maximum numbers of parameters and FLOPs are:
$\tau_{min}^{PARAMS}=0.1 M$, $\tau_{max}^{PARAMS}=2.0 M$, $\tau_{min}^{FLOPs}=0.02 B$, $\tau_{max}^{FLOPs}=0.31 B$.
Despite opposite claims in the literature \cite{lu2018nsga}, we found it simple to calculate the various targets because information about the operators in the computational graph is readily available (e.g.,  number of parameters).

\noindent\textbf{Pre-trained shared model.} 
Our shared model stacks 6 normal and 2 reduction cells. The reduction cells are placed in the 3rd and 6th cell. 
We choose 1270 epochs for the pre-training phase, by measuring when the performance on batches of 64 randomly sampled architectures stabilizes. 

\noindent\textbf{Controller.} The controller is designed as an LSTM with hidden size 64. For each layer, the controller samples an operation and an input node from previous nodes. For ADF (resp., ADC), the controller is trained with Adam with a learning rate of 0.001 (resp., 0.002). A tanh constant 1.5 is used for sampling logits to prevent premature convergence \cite{tanh}. The temperature varies from 10 to 5 (resp., from 25 to 1) with a period of 50 (resp., 1200) for ADF (resp., ADC).
The $\epsilon$ for the density function in \eqref{eq:reward} is set to $0.1 M$ and $0.02 B$ for the number of parameters and FLOPs, respectively. The exploration hyperparameters are determined by grid search on CIFAR-10, but then reused on CIFAR-100.

\noindent\textbf{Search methods.} We choose random search (RS) and desirability function with multiple runs (M-DF) as baselines. The former randomly samples child models. For two objectives, the latter first splits the interval $[\tau^{PARAM}_{min},\tau^{PARAM}_{max}]$ into equal parts using 10 split points in total. Subsequently, we do 10 independent runs using reward function \eqref{eq:DF-reward}. During each run, the target of the desirability function is fixed to one of the split points so that the controller focuses on child models near the split point. For three objectives, M-DF creates a regular 4 by 4 grid to split the 2-D interval $[\tau^{PARAM}_{min},\tau^{PARAM}_{max}]\times[\tau^{FLOPs}_{min},\tau^{FLOPs}_{max}]$, and uses the 16 grid points as targets for 16 independent runs. 
Each of the independent runs of M-DF uses the same number of search steps as RS and NPG-NAS to ensure that M-DF can converge to a good optimal policy for the corresponding target.
M-DF serves as our approximate ground truth. Due to the immense search space, the exact Pareto front is beyond reach. M-DF thus acts as a strong baseline that estimates different parts of the Pareto front, and subsequently combines them to approximate the full front.
This is why we allocate it considerably more search steps than other methods.

\noindent\noindent\textbf{Hyperparameters for search methods.}
For two objectives, we use 6000 search steps for RS, ADC, and ADF so that all methods have the same budget and comparison is fair. ADF uses 1500 additional steps for the warm-up phase.  
For M-DF, we sample 6000 models in each run for a total of 60000 samples.
M-DF uses the same hyperparameters for the desirability function as ADF except that it uses a fixed temperature of 5 (like ENAS) and does not execute the warm-up phase.
For three objectives, we double the search steps for RS and ADC to 12000. 
ADF uses target pairs consisting of the number of parameters and number of FLOPs.
To perform annealing, we split the 2-D interval $[\tau^{PARAM}_{min},\tau^{PARAM}_{max}]\times[\tau^{FLOPs}_{min},\tau^{FLOPs}_{max}]$ using a 110 by 109 regular grid of 11990 points. This results in almost the same number of steps as RS and ADC. During the annealing phase, we traverse the grid in a zig-zag manner (i.e., in diagonal strips) to ensure that any two consecutive target pairs are sufficiently close and policy transfer is possible.
The desirability function is taken as the product of two 1-D desirability functions, one for number of parameters and another one for number of FLOPs. 
For M-DF, we sample 6000 models in each run for a total of 6000x16=96000 search steps.

\noindent\textbf{Hyperparameters for training from scratch.}
To train from scratch selected architectures on CIFAR-10 and CIFAR-100, we use CutOut~\cite{devries2017improved} and random horizontal flip for data augmentation. The optimizer is SGD with cosine decay learning rate 0.025, weight decay 0.0003, Nesterov momentum 0.9. Each model is trained for 600 epochs.
To train from scratch on ImageNet, we use SGD with cosine learning rate 0.8, weight decay 0.00003, and Nesterov momentum 0.9. Each model is trained for 350 epochs.

\begin{figure}[t!]
	\centering
	\begin{subfigure}[t]{0.2\textwidth}
		\centering
		\includegraphics[width=\textwidth]{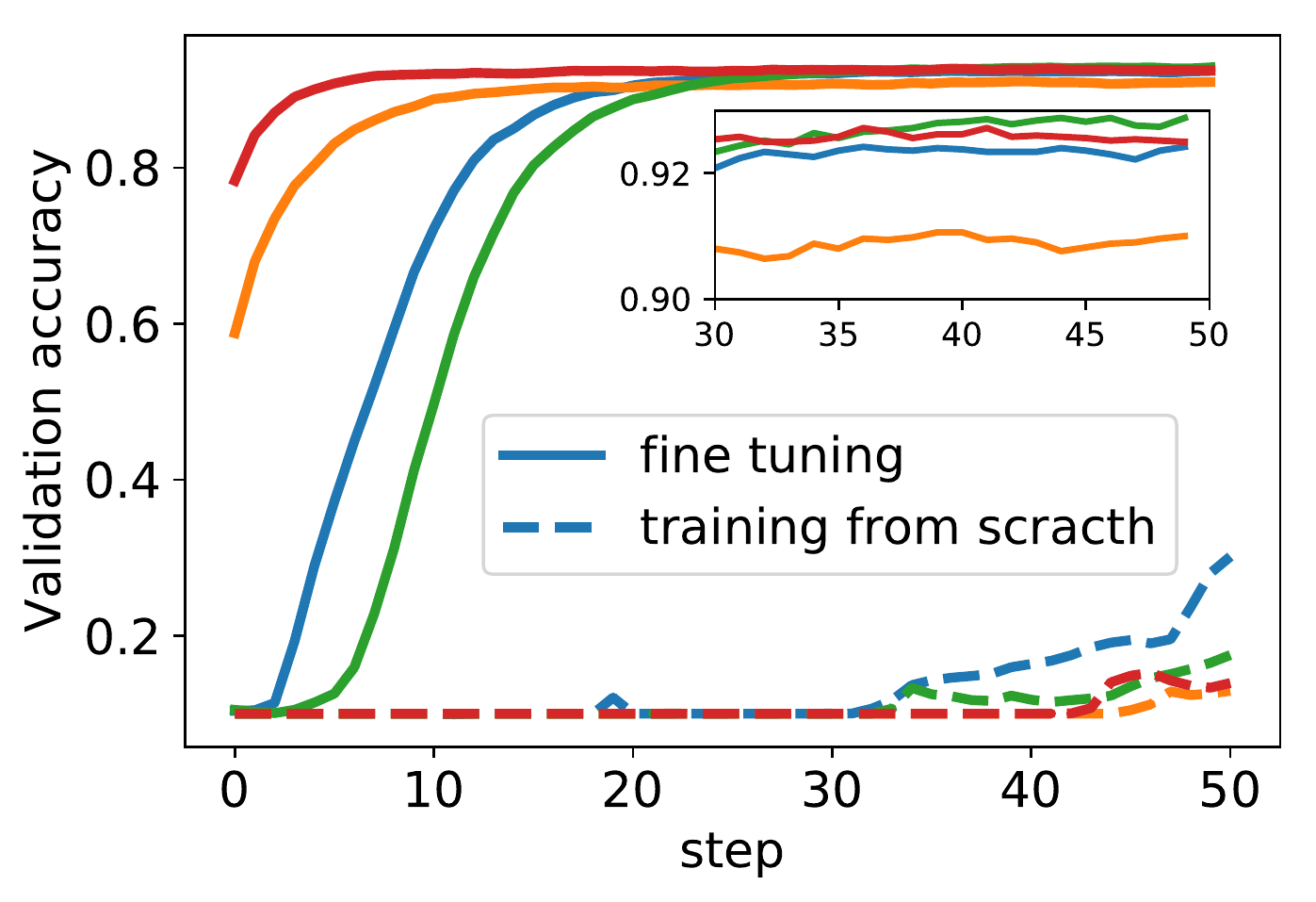}
		\caption{}
	\label{fig:finetune}
	\end{subfigure}
	\begin{subfigure}[t]{0.2\textwidth}
		\centering
		\includegraphics[width=\textwidth]{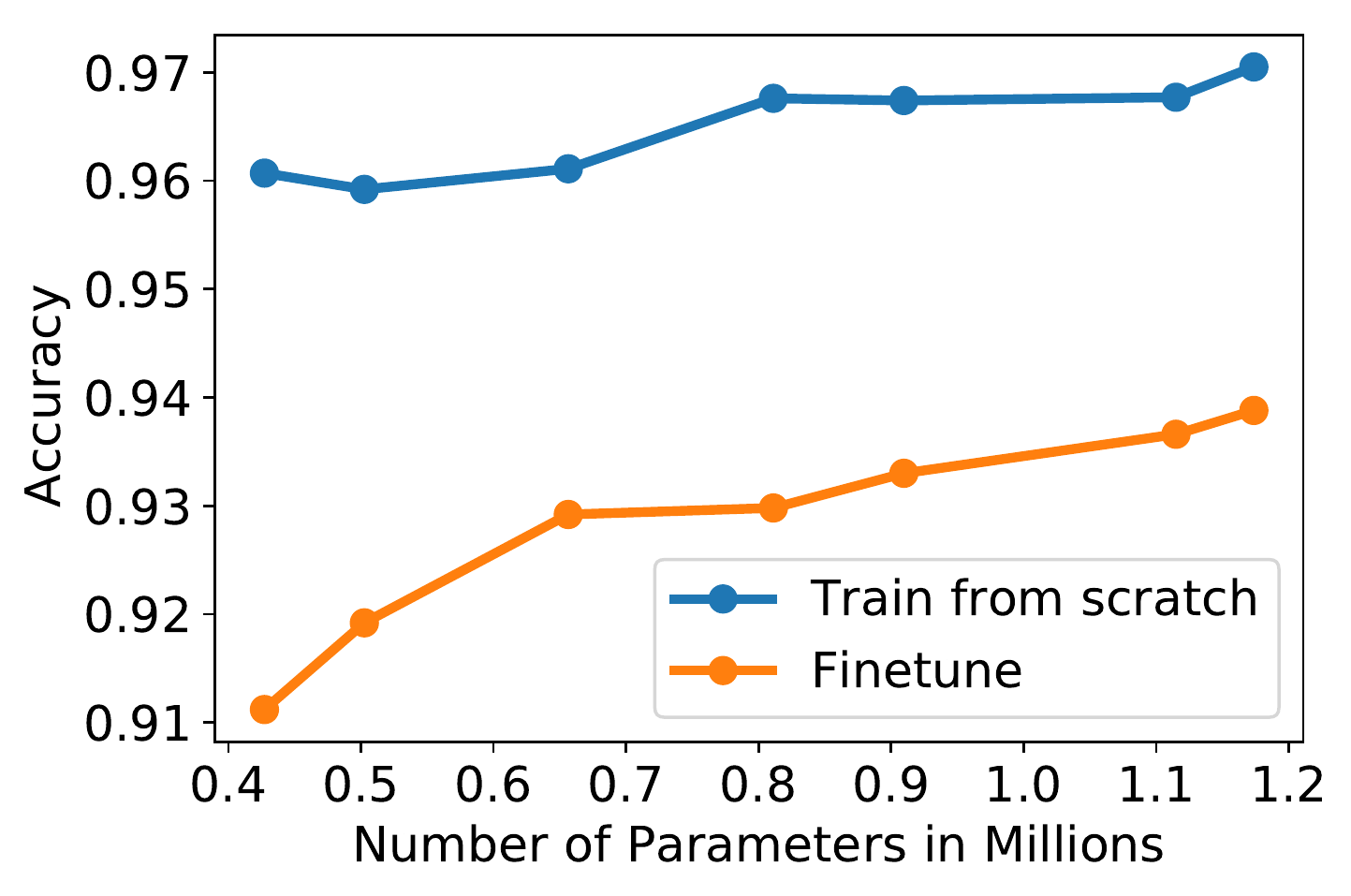}
		\caption{}
	\end{subfigure}
	\caption{(a) Validation accuracy when fine-tuning shared model and training from scratch with random initialization vs. training step. The 4 colors represent 4 random architectures. 
	(b) Performance of different child models during search and final evaluation phase.}
	\label{fig:train_from_scratch}
\end{figure}

\subsection{Results}\label{sec:exp-results}
\noindent\textbf{Effectiveness of pre-trained shared model.}
In Fig.~\ref{fig:train_from_scratch}(a) we show the validation accuracy as a function of training step for 4 random architectures that are trained from scratch (with random initialization) compared with fine-tuning from our pre-trained shared model. Remarkably, the validation accuracy by fine-tuning converges quickly within 30 training steps.
We speculate that the pre-training phase learns weights for the operators of the shared model that are able to extract good features (in the convolutional layers), so that a few gradient steps suffice for any child model to converge.
We use 50 steps for guaranteed convergence. Second, we pick 7 representative architectures with different numbers of parameters from a searched Pareto front on CIFAR-10 with ADC (see experiments below), and train them from scratch with random initialization. Fig.~\ref{fig:train_from_scratch}(b) shows the test accuracy with random initialization and the validation accuracy produced by fine-tuning. The two are strongly correlated at a Pearson correlation coefficient of 0.84. This implies an effective ranking of the architectures' performance, for all parts of the Pareto front. As argued in \cite{pmlr-v80-bender18a}, the high correlation is possibly because weight sharing implicitly forces the one-shot model to identify and focus on the operations that are most useful for generating good predictions. Architectures with high accuracy from fine-tuning will then also have high quality with stand-alone training.

\begin{figure}[t!]
    \centering
    \includegraphics[scale=0.33]{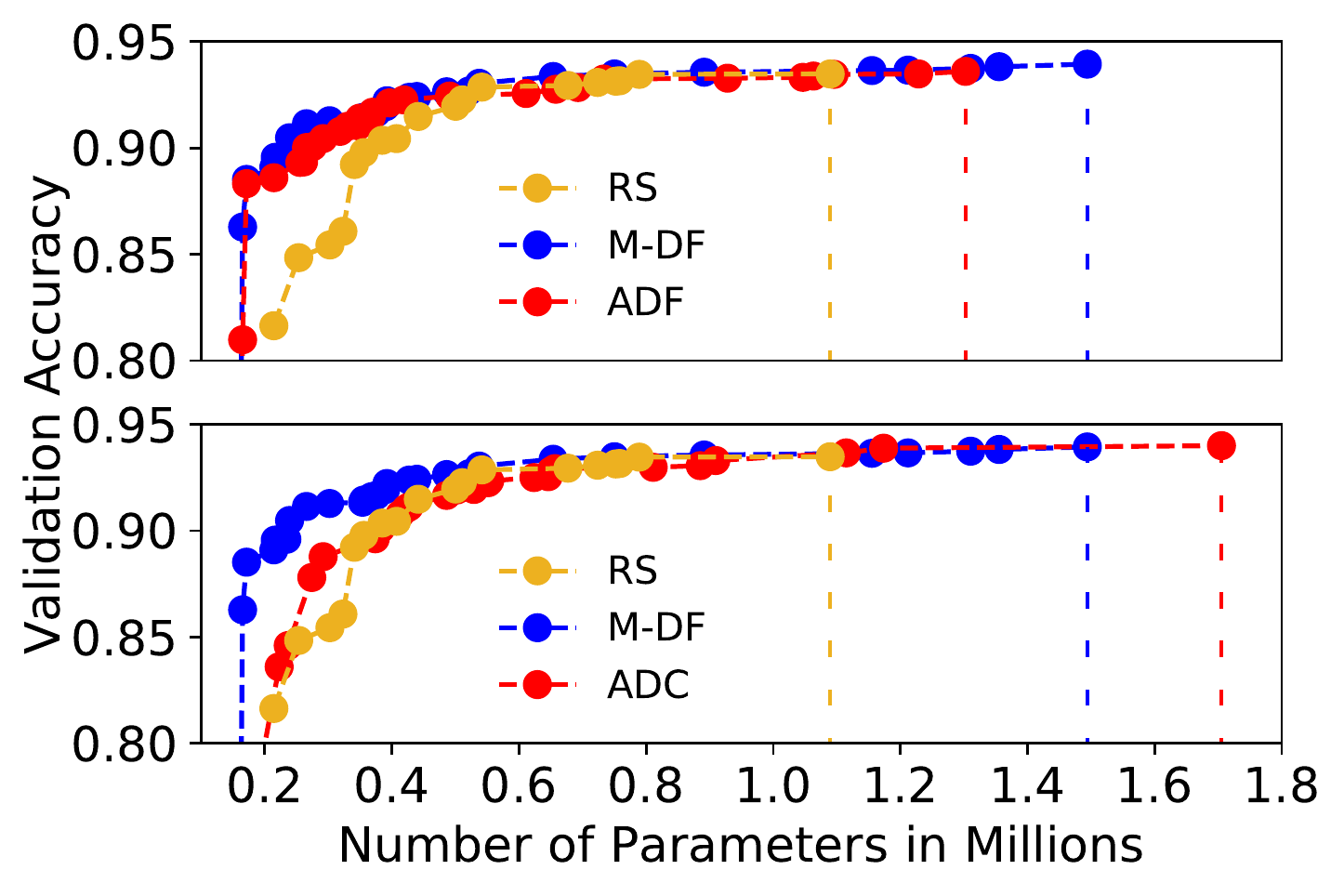}
\caption{Pareto fronts for validation accuracy and number of parameters obtained by ADF (up), ADC (bottom), RS, and M-DF on CIFAR-10. Dashed lines indicate termination of Pareto fronts.}
    \label{fig:flt64_all}
\end{figure}

\begin{figure}[t!]
    \centering
    \includegraphics[scale=0.29]{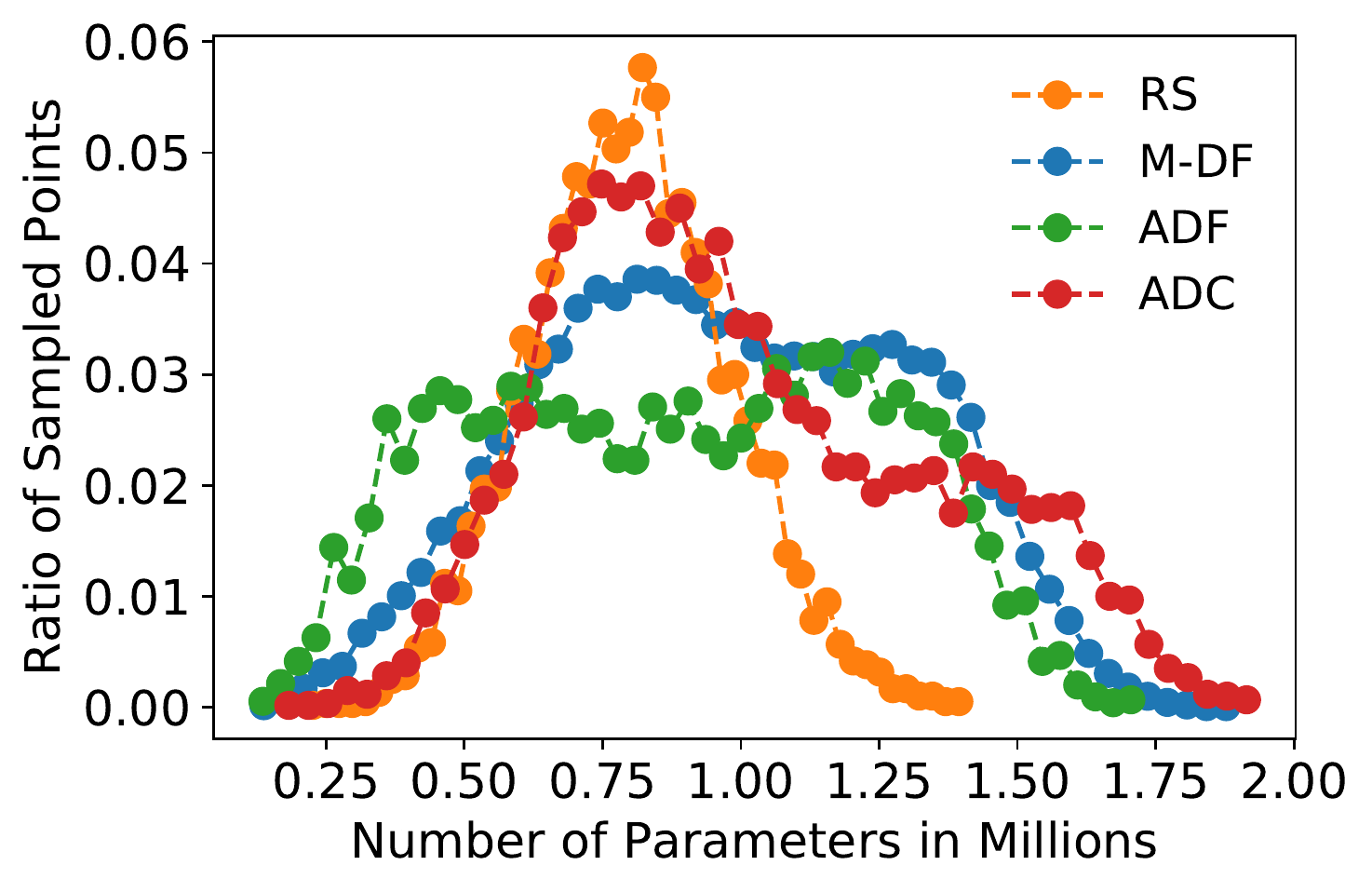}
\caption{Histograms for different methods.}
    \label{fig:sample_distribution_all}
\end{figure}

\noindent\textbf{Two objectives on CIFAR-10. } Fig.~\ref{fig:flt64_all} shows that
ADF and ADC perform similarly to M-DF and approximate the full front while significantly outperforming RS, which retrieves a small part only. 
The poor performance of RS is in stark contrast to its strong performance in single-objective NAS~\cite{yang2019nas}.
For more insight, we plot the child model distributions (histograms) for all methods in Fig.~\ref{fig:sample_distribution_all}. RS only samples a small part of the search space while all other methods sample the search space more uniformly.
ADF achieves the most uniform sampling, even compared to M-DF, because it expands the Pareto front from the leftmost to the rightmost target at a constant pace.
ADC also explores a larger subspace than RS. It extends the Pareto front to the rightmost limit, despite performing worse than ADF in the leftmost region.
Appendix \ref{appendix:scatter_plots} provides more details on the child model distributions.

\begin{figure}[t!]
    \centering
    \includegraphics[scale=0.33]{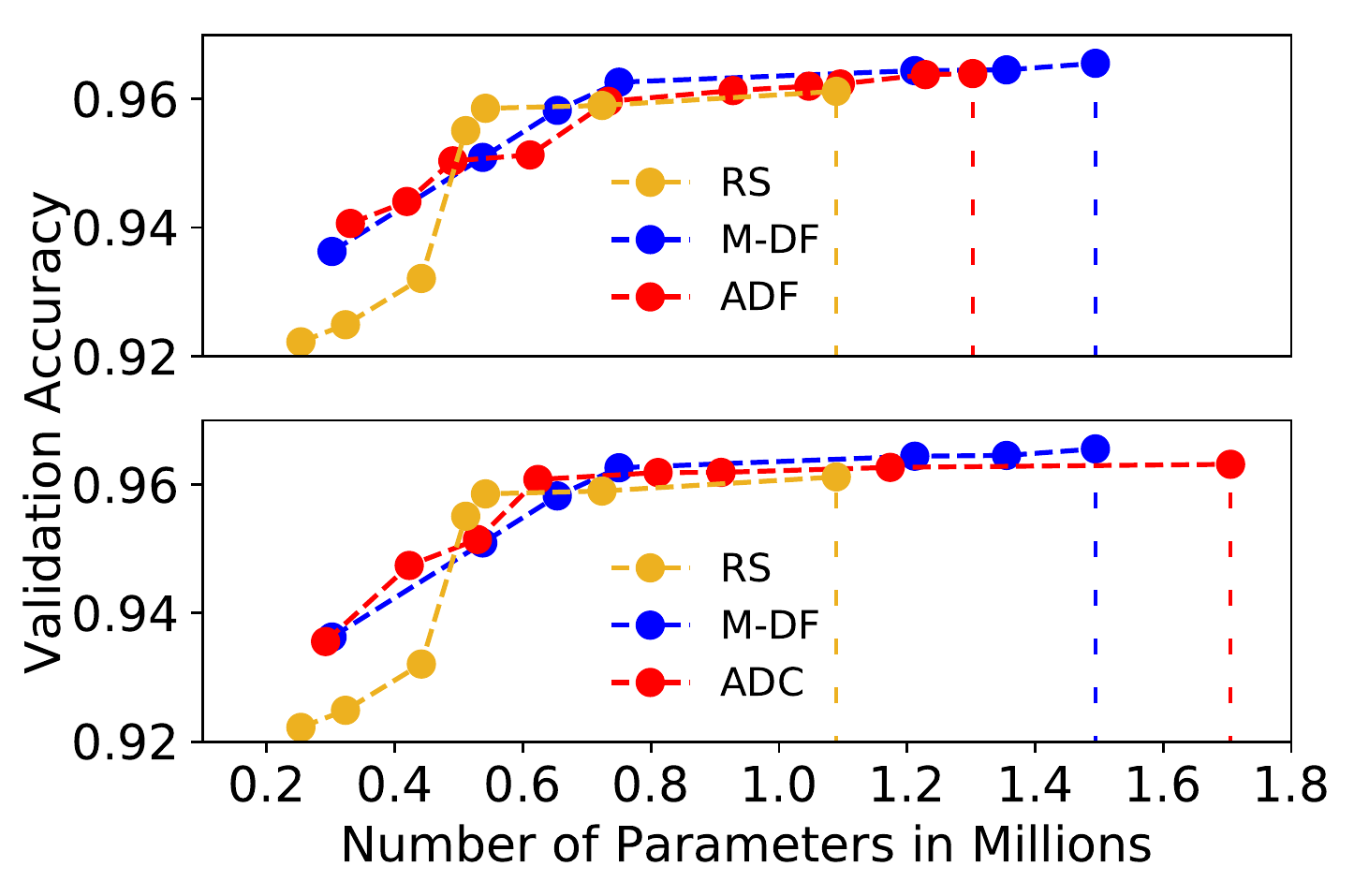}
    \caption{Pareto fronts of Fig. \ref{fig:flt64_all} trained from scratch.}
    \label{fig:pareto_from_scratch}
\end{figure}

The Pareto fronts in Fig.~\ref{fig:flt64_all} are produced by continuously fine-tuning the shared model using 50 gradient steps per sampled architecture. A critical question concerns the form of these curves when architectures are trained from scratch. To answer this, we pick 10 evenly distributed points from each curve, and train them from scratch with random initialization. The corresponding Pareto fronts are depicted in Fig. \ref{fig:pareto_from_scratch}. We again observe that our methods perform similarly to M-DF while outperforming RS.
This is hardly surprising given the effectiveness of our pre-trained shared model. 
RS, in particular, performs badly in the leftmost region, while being unable to extend the Pareto front to the right. It only outperforms other methods in a small oversampled subspace.

NPG-NAS is an efficient multi-objective NAS framework. On a single Nvidia Tesla P100 GPU, the search phase of ADF/ADC takes 75/60 GPU hours respectively, while M-DF takes 605 GPU hours. The search cost for ADF or ADC is lower than the 4/8 GPU days for NSGA-Net-micro/NSGA-Net-macro based on a micro/macro search space (see also Table~\ref{table:result}). The total runtime for ADF/ADC is 7.25/6.63 GPU days though due to 99  additional GPU hours to pre-train the shared model. This is higher than NSGA-Net-micro but lower than NSGA-Net-macro. However, the reason behind the high efficiency of NSGA-Net is that it trains sampled architectures for only 25 epochs (instead of 600 epochs that they use for full training). Early stopping is nevertheless a low fidelity estimate with high bias~\cite{survey}, and no evidence is provided that the accuracies from 25 epochs are highly correlated with the fully trained ones, for different parts of the Pareto front. This can harm the ability to approximate the true Pareto front well.
Our efficiency becomes more pronounced when we compare to LEMONADE, which evaluates child models based on the accurate network morphisms~\cite{Wei:2016:NM:3045390.3045451}.
LEMONADE takes 1344 GPU hours on Nvidia Titan X GPU in total; ADF and ADC are 7.7x and 8.5x faster.

\begin{table}[t!]    
\tiny
    \centering
    \caption{Areas dominated by the Pareto fronts averaged over 5 runs with standard deviation of RS, M-DF and NPG-NAS.}
    \label{table:area}
    \begin{adjustbox}{max width=.45\textwidth}
        \begin{tabular}{|c|c|c|c|c|}
\hline
    & \multicolumn{2}{c}{fixed temperature}  & \multicolumn{2}{|c|}{cosine temperature} \\ \cline{2-5}
Method   & Mean & SD & Mean & SD\\ 
\hline
RS    & 0.826 & 0.0068 & - & - \\
\hline
M-DF  & 0.876 & 0.0026 & - & - \\
\hline
M-DF (2/3 targets)  & 0.817/0.824 & 0.0188/0.0075 & - & - \\
\hline
ADF   & 0.873 & 0.0053 & 0.873 & 0.0028 \\
\hline
ADC   & 0.805 & 0.0404 & 0.872 & 0.0114\\
\hline
        \end{tabular}
    \end{adjustbox}
\end{table}

We further assess the quality of the Pareto fronts using the area they dominate, with the number of parameters rescaled to $[0,1]$. 
Due to the high GPU hour cost, we only do 5 runs for each method. 
NPG-NAS incorporates by default our exploration scheme. We repeat our methods with a fixed temperature $T$=5, like ENAS.
The mean and standard deviation from all runs are summarized in Table \ref{table:area}. 
NPG-NAS outperforms RS, while performing similarly to M-DF. ADF has lower variance and thus more stable performance than RS, whereas ADC has visibly larger variance.
We find that the temperature scheme has a negligible effect on the mean of ADF, but it cuts the variance by almost 50\%. For ADC, the varying temperature may play a more critical role. It increases the mean but also reduces the variance. In particular, without it the performance of ADC is even worse than RS. Our scheme thus seems to have the potential to increase or stabilize performance under non-stationary rewards.
Finally, to compare against M-DF with same budget, we run M-DF with 2 (3) targets, each with 3000 (2000) search steps. We get a mean of 0.817 (0.824), which are worse than even RS.

\begin{figure}[t!]
    \centering
    \includegraphics[scale=0.29]{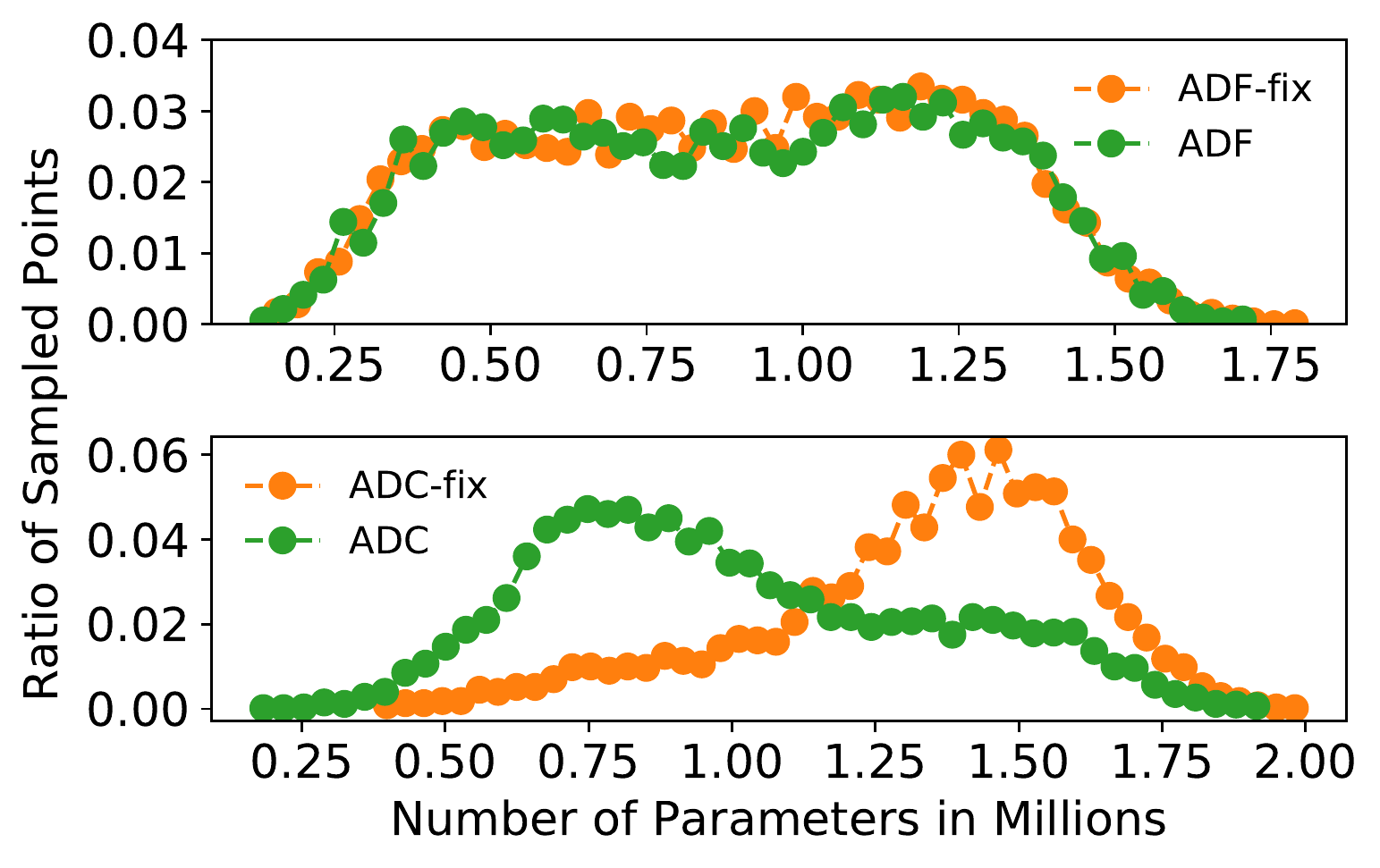}
    \caption{Comparison between ADF (top) and ADC (bottom) with cosine temperature decay and with fixed $T=5$.}
        \label{fig:cos_temp}
\end{figure}

To gain more insight into the results in Table~\ref{table:area}, we depict in Fig.~\ref{fig:cos_temp} the distribution of sampled architectures for ADF and ADC in the two cases. The two histograms are similar for ADF, but dramatically different for ADC. Without temperature decay, ADC oversamples a small subspace centered around 1.4M parameters.
We conjecture that cosine temperature decay helps it escape this subspace, by encouraging the controller to randomly sample other parts of the search space. ADF does not seem to suffer from this problem. Still, improved exploration in ADF may help to reduce the effect from previous policies, which is a possible reason for the variance reduction.

\begin{table}[t!]    
    \centering
    \caption{Classification accuracies on CIFAR-10. Blocks from top to bottom present architectures designed by: experts; other NAS methods; other multi-objective NAS methods; NPG-NAS (ADF/ADC).}
    \label{table:result}
    \begin{adjustbox}{max width=.47\textwidth}
    \begin{tabular}{c c c c}
        \toprule
        \toprule
        Method& GPU (days) & Params (million) & accu(\%)\\
        \midrule
        DenseNet + cutout~\cite{devries2017improved} & -   & 26.2 & 97.44\\
        \midrule
        \midrule
        NAS~\cite{zoph2017neural} & 22400 & 7.1 & 95.53\\
        SMASH~\cite{brock2018smash} & 1.5 & 16.0 & 95.97\\
        ENAS-L/M/S + micro + cutout~\cite{pham2018efficient} & 0.45 & 4.6/2.82/1.27 & 97.11/96.87/96.49\\
        NASNet-A + cutout~\cite{zoph2018learning} & 2000 & 3.3 & 97.35\\
        DARTS (second order) + cutout~\cite{liu2018darts} & 4 & 3.3 & 97.24 \\
        PNAS~\cite{Liu_2018_ECCV} & 225 & 3.2 & 96.59 \\
        AmoebaNet-B (N=6, F=36) + cutout~\cite{Real_Aggarwal_Huang_Le_2019} & 3150 & 2.8 & 97.45 \\
         \midrule
         \midrule
         NSGA-Net-microv1~\cite{lu2018nsga}& 4 & 3.3  & 97.25 \\
         NSGA-Net-microv2~\cite{lu2018nsga}& 4 & 26.8 & 97.50 \\
         NSGA-Net-macro~\cite{lu2018nsga}& 8 & 3.3 & 96.15 \\
         DPP-Net-PNAS~\cite{dong2018dpp} & 8 & 11.39 & 95.64 \\
         LEMONADE~\cite{elsken2018efficient} & 56 & 13.1 & 97.42 \\
         LEMONADE~\cite{elsken2018efficient} & 56 & 4.7 & 96.95 \\
         \midrule
         \midrule
         ADF-L/M/S & 7.25 (total cost) & 3.94/2.94/1.18 & 97.24/96.98/96.90\\
         ADC-L/M/S & 6.63 (total cost) & 4.13/2.77/1.56 & 97.52/97.24/96.72\\
        \bottomrule
        \bottomrule
    \end{tabular}
    \end{adjustbox}
\end{table}

To evaluate the found cells with respect to state-of-the-art methods, we pick three representative cells with large (L), medium (M) and small (S) number of parameters from the Pareto fronts of ADF and ADC, and evaluate them on CIFAR-10 with 36 channels and 20 cells. 
This is important because cells are typically searched on small neural nets for efficiency purposes, but then stacked at higher depths and with more filters (width) for higher predictive performance.
Furthermore, we train the ENAS cell by stacking it 20, 14, or 8 times (ENAS-L/M/S, respectively)\footnote{ENAS-L is reported in \cite{pham2018efficient}. ENAS-M/S are evaluated using the ENAS open-source code.}.
Table~\ref{table:result} summarizes results from manually designed architectures, NPG-NAS and other NAS methods. We do not depict prior results using data augmentation techniques except for CutOut.

NPG-NAS is consistently better than ENAS-L/M/S, showing that searching for the entire Pareto front can provide performance benefit compared to stacking a fixed cell at varying depths.
ADC-L outperforms the expert-designed DenseNet with only about 16\% of its parameters, and outperforms many other NAS algorithms. Compared to other multi-objective algorithms such as LEMONADE and DPP-Net-PNAS, NPG-NAS has substantially lower GPU cost while obtaining higher accuracy with fewer parameters. 
All our cells have higher accuracy than NSGA-Net-macro, including ADF-S at 1.18M parameters.
Our cells ADF-L and ADC-M perform almost identically to NSGA-Net-microv1, even though only the latter has fewer parameters.
On the other hand, ADC-L achieves higher accuracy than NSGA-Net-microv2 with just 15.41\% its parameters.

Even though a direct comparison of different cells or architectures is tricky due to the different search spaces, 
the results in 
Table~\ref{table:result} point to the high quality of our discovered cells. This is hardly surprising, given the strong correlation between the fine-tuned accuracies vs. the stand-alone ones.

\begin{figure}[t!]
    \centering
    \includegraphics[scale=0.33]{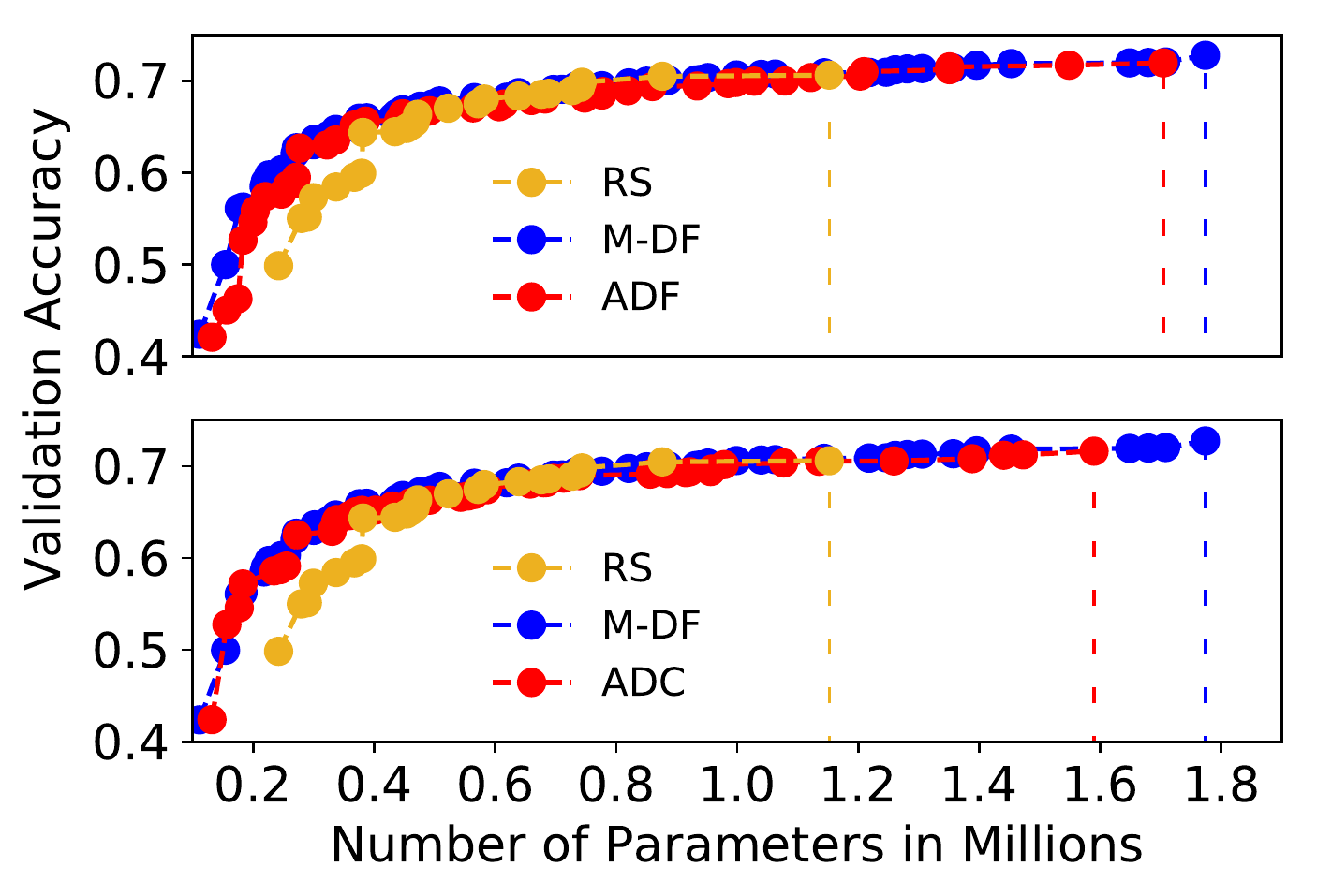}
    \caption{Pareto fronts on CIFAR-100.}
    \label{fig:flt64_100_all}
\end{figure}

\noindent\textbf{Two objectives on CIFAR-100.} NPG-NAS is stable across different datasets. 
To illustrate this, we experiment with dataset CIFAR-100 using exactly the same exploration hyperparameters (e.g., $T_{min},T_{max},\nu$) as for CIFAR-10.
Having transferable hyperparameters helps to avoid the cost associated with optimizing the hyperparameters for a new dataset from scratch, and is thus very beneficial.
The results are shown in Fig.~\ref{fig:flt64_100_all}. Our methods outperform RS and yield competitive results to M-DF. 
We also evaluate the performance of the found architectures. The results are shown in Table~\ref{table:result_100}. ADF-L outperforms DenseNet-BC with 77.61\% fewer parameters. It is only slightly lower than P-DARTSv1. P-DARTSv2 and NAONet outperform our cell by larger margins, but their size is also much bigger. Compared to the transferred cell of NSGA-Net, NPG-NAS outperforms it by a large margin with fewer parameters and lower search cost.

\begin{table}[t!]    
    \centering
    \caption{Classification accuracies on CIFAR-100.}
    \label{table:result_100}
    \begin{adjustbox}{max width=.47\textwidth}
        \begin{tabular}{c c c c}
            \toprule
            \toprule
            Method& GPU (days) & Params (million) & accu(\%)\\
            \midrule
            DenseNet-BC ($k$=40)~\cite{8099726} & -   & 25.6 & 82.82\\
            ResNeXt + shake-shake + cutout~\cite{devries2017improved} & -   & 34.4 & 84.80\\
            \midrule
            \midrule           MetaQNN~\cite{baker2017designing} & 100 & 11.18 & 72.86\\
            SMASHv1~\cite{brock2018smash} & - & 4.6 & 77.93\\
            SMASHv2~\cite{brock2018smash} & 3.0 & 16.0 & 79.4\\
            P-DARTSv1~\cite{pdarts} & 0.3 & 3.6 & 84.08 \\
            P-DARTSv2~\cite{pdarts} & 0.3 & 11.0 & 85.36 \\
            NAONet + cutout~\cite{NAO} & 200 & 10.8 & 84.33\\
            \midrule
            \midrule
            NSGA-Netv1~\cite{lu2018nsga}& 8 & 3.3  & 79.26 \\
            NSGA-Netv2~\cite{lu2018nsga}& 8 & 11.6  & 80.17 \\
            \midrule
            \midrule
            ADF-L/M/S & 6.71 (total cost) & 5.73/4.27/1.91 & 83.99/82.67/80.76\\
            ADC-L/M/S & 6.08 (total cost) & 5.38/3.47/0.90 & 83.19/82.18/77.8\\
            \bottomrule
            \bottomrule
        \end{tabular}
    \end{adjustbox}
\end{table}

\noindent\textbf{Transfer to ImageNet.}
The large cells found from CIFAR-10 and CIFAR-100 are stacked with 48 channels and 14 layers. An SE \cite{hu2018squeeze} module is added after each output node. The results are summarized in Table \ref{table:ImageNet}. ADF-L10 achieves higher top-1 accuracy than MobileNetv2-1.4 with 19.42\% fewer parameters.
NPG-NAS beats most other methods at similar network sizes. ADC-L10 and ADF-L100 get higher accuracy than DPP-Net-PNAS with only 1/10 of its parameters. Furthermore, we outperform other NAS methods by clear margins at similar network sizes. Even though MnasNet-A3 has the highest accuracy, it consumes a large amount of GPU resources (about 91000 GPU hours estimated by FBNet) to search directly on ImageNet.

\begin{table}[ht!]    
    \centering
    \caption{Classification accuracies on ImageNet. 10/100 denote cell transfer from CIFAR-10/CIFAR-100.}
    \label{table:ImageNet}
    \begin{adjustbox}{max width=.45\textwidth}  
        \begin{tabular}{c c c c}
            \toprule
            \toprule
            Method& Params (million) & top-1 accu(\%) & top-5 accu(\%)\\
            \midrule
            MobileNetv2~\cite{8578572} & 3.4  & 72.0 & 91.0\\
            MobileNetv2-1.4~\cite{8578572} & 6.9  & 74.7 & 92.5\\
            \midrule
            \midrule
            NASNet-A~\cite{zoph2018learning} & 5.3 & 74.0 & 91.6\\
             DARTS~\cite{liu2018darts}~ & 4.7 & 73.3 & 91.3\\
            PNASNet~\cite{Liu_2018_ECCV}~ & 5.1 & 74.2 & 91.9\\
            \midrule
            \midrule
            DPP-Net-PNAS~\cite{dong2018dpp}& 77.16 & 75.84  & 92.87 \\
            DPP-Net-Panacea~\cite{dong2018dpp}& 4.8 & 74.02  & 91.79 \\
            MnasNet-A3~\cite{tan2018mnasnet}& 5.2 & 76.7  & 93.3 \\
            LEMONADE~\cite{elsken2018efficient} & 6.0 & 71.7 & 90.4 \\
            FBNet-C~\cite{wu2018fbnet} & 5.5 & 74.9 & - \\
            \midrule
            \midrule
            ADF/C-L10 &  5.56/7.24 & 75.36/76.66 & 92.39/93.03 \\
            ADF/C-L100 & 7.89/7.43 & 76.67/76.61 & 92.90/92.82 \\
            \bottomrule
            \bottomrule
        \end{tabular}
    \end{adjustbox}
\end{table}

\begin{figure}[t!]
    \centering
    \includegraphics[width=0.26\textwidth]{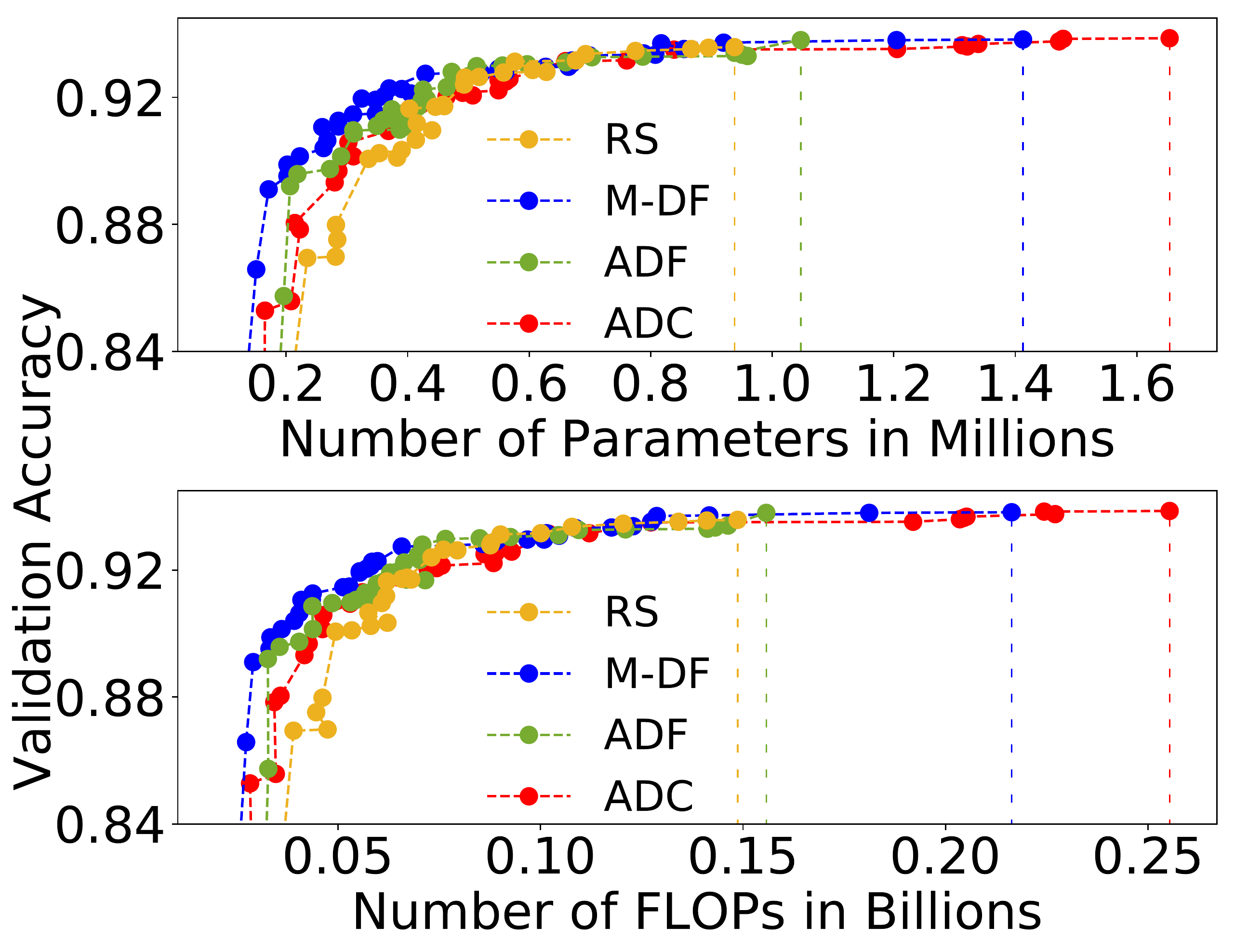}
    \caption{Projected Pareto fronts  for validation accuracy vs number of parameters (up) and validation accuracy vs. FLOPs (bottom) obtained by ADF, ADC, M-DF and RS.} 
    \label{fig:three_objective_cifar10}
\end{figure}

\noindent\textbf{Three objectives on CIFAR-10. }
Finally, we experiment with three objectives in Fig. \ref{fig:three_objective_cifar10}. 
Both ADF and ADC visibly outperform RS and perform quite similarly to M-DF. 
We observe that ADF performs better in the leftmost regions in Fig.~\ref{fig:three_objective_cifar10} while its top accuracy in the rightmost region just a bit lower than M-DF and ADC.
ADC achieves a slightly higher range of accuracies, and has a more extended right front.

\subsection{Discussion}
In multi-objective NAS, the true Pareto front is unattainable due to the huge search space. Our goal is an approximation with the following properties: accuracy, good range, and diversity \cite{Laumanns:2002,1197687}.
Accuracy asks whether we converge to the true Pareto front. Both ADF and ADC generally produce Pareto fronts similar to M-DF, which serves as our (approximate) ground truth. 
This provides positive evidence in favor of convergence. 
Note that our pre-trained shared model contributes to accuracy, because the fine-tuned accuracies are strongly correlated with the stand-alone ones.
We further note that ADF and ADC sample the search space much more uniformly than RS. 
For this reason, our methods also have better range, and produce a more extended Pareto front than RS (ADC even more so than ADF).
Diversity also holds, since NPG-NAS produces diverse solutions during search. 
Diversity is visualized in Appendix \ref{appendix:discovered_cells}.
\section{Conclusion}
NPG-NAS is based on the new paradigm of non-stationary policy gradient. 
It is a flexible and generic framework: depending on the reward function definition, it can accommodate very different algorithms that incorporate elements from the dominant paradigms of scalarization and evolution.
Experiments on many datasets show that NPG can be an efficient and effective paradigm for multi-objective NAS.

Future research can shed more light on its fundamentals. A major question concerns the better theoretical understanding of the non-stationary policy adaptation mechanism for ADF and ADC. 
A second practical question concerns other non-stationary reward functions within NPG-NAS with possibly different underlying mechanisms of policy adaptation.
Finally, it is critical to gain insights into the impact of our proposed exploration scheme on the performance of reinforcement learning, and to even study alternative schemes for non-stationary reward functions. 
\bibliography{main}

\begin{thebibliography}{}

\bibitem[\protect\citeauthoryear{Al{-}Shedivat \bgroup et al\mbox.\egroup
  }{2018}]{Al-ShedivatBBSM18}
Al{-}Shedivat, M.; Bansal, T.; Burda, Y.; Sutskever, I.; Mordatch, I.; and
  Abbeel, P.
\newblock 2018.
\newblock Continuous adaptation via meta-learning in nonstationary and
  competitive environments.
\newblock In {\em ICLR}.

\bibitem[\protect\citeauthoryear{Baker \bgroup et al\mbox.\egroup
  }{2017}]{baker2017designing}
Baker, B.; Gupta, O.; Naik, N.; and Raskar, R.
\newblock 2017.
\newblock Designing neural network architectures using reinforcement learning.
\newblock In {\em ICLR}.

\bibitem[\protect\citeauthoryear{Bello \bgroup et al\mbox.\egroup
  }{2017a}]{tanh}
Bello, I.; Pham, H.; Le, Q.~V.; Norouzi, M.; and Bengio, S.
\newblock 2017a.
\newblock Neural combinatorial optimization with reinforcement learning.
\newblock In {\em ICLR workshop}.

\bibitem[\protect\citeauthoryear{Bender \bgroup et al\mbox.\egroup
  }{2018}]{pmlr-v80-bender18a}
Bender, G.; Kindermans, P.-J.; Zoph, B.; Vasudevan, V.; and Le, Q.
\newblock 2018.
\newblock Understanding and simplifying one-shot architecture search.
\newblock In {\em ICML}.

\bibitem[\protect\citeauthoryear{Besbes, Gur, and Zeevi}{2014}]{Besbes:2014}
Besbes, O.; Gur, Y.; and Zeevi, A.
\newblock 2014.
\newblock Stochastic multi-armed-bandit problem with non-stationary rewards.
\newblock In {\em NeurIPS}.

\bibitem[\protect\citeauthoryear{Brock \bgroup et al\mbox.\egroup
  }{2018}]{brock2018smash}
Brock, A.; Lim, T.; Ritchie, J.; and Weston, N.
\newblock 2018.
\newblock {SMASH}: One-shot model architecture search through hypernetworks.
\newblock In {\em ICLR}.

\bibitem[\protect\citeauthoryear{Cai, Zhu, and Han}{2019}]{cai2018proxylessnas}
Cai, H.; Zhu, L.; and Han, S.
\newblock 2019.
\newblock Proxyless{NAS}: Direct neural architecture search on target task and
  hardware.
\newblock In {\em Int. Conf. Learning Representations}.

\bibitem[\protect\citeauthoryear{Chen \bgroup et al\mbox.\egroup
  }{2019}]{pdarts}
Chen, X.; Xie, L.; Wu, J.; and Tian, Q.
\newblock 2019.
\newblock Progressive differentiable architecture search: Bridging the depth
  gap between search and evaluation.
\newblock In {\em IEEE ICCV}.

\bibitem[\protect\citeauthoryear{da Silva \bgroup et al\mbox.\egroup
  }{2006}]{daSilva:2006}
da~Silva, B.~C.; Basso, E.~W.; Bazzan, A. L.~C.; and Engel, P.~M.
\newblock 2006.
\newblock Dealing with non-stationary environments using context detection.
\newblock In {\em Proc. Int. Conf. Machine Learning}.

\bibitem[\protect\citeauthoryear{Deb \bgroup et al\mbox.\egroup
  }{2000}]{deb2000fast}
Deb, K.; Agrawal, S.; Pratap, A.; and Meyarivan, T.
\newblock 2000.
\newblock A fast elitist non-dominated sorting genetic algorithm for
  multi-objective optimization: {NSGA-II}.
\newblock In {\em PPSN}.

\bibitem[\protect\citeauthoryear{Derringer and
  Suich}{1980}]{derringer1980simultaneous}
Derringer, G., and Suich, R.
\newblock 1980.
\newblock Simultaneous optimization of several response variables.
\newblock {\em J. of Quality Technology} 12(4):214--219.

\bibitem[\protect\citeauthoryear{DeVries and
  Taylor}{2017}]{devries2017improved}
DeVries, T., and Taylor, G.~W.
\newblock 2017.
\newblock Improved regularization of convolutional neural networks with cutout.
\newblock {\em arXiv preprint arXiv:1708.04552}.

\bibitem[\protect\citeauthoryear{Dong \bgroup et al\mbox.\egroup
  }{2018}]{dong2018dpp}
Dong, J.-D.; Cheng, A.-C.; Juan, D.-C.; Wei, W.; and Sun, M.
\newblock 2018.
\newblock Dpp-net: Device-aware progressive search for pareto-optimal neural
  architectures.
\newblock In {\em Eur. Conf. Computer Vision}.

\bibitem[\protect\citeauthoryear{Elsken, Metzen, and
  Hutter}{2019a}]{elsken2018efficient}
Elsken, T.; Metzen, J.~H.; and Hutter, F.
\newblock 2019a.
\newblock Efficient multi-objective neural architecture search via lamarckian
  evolution.
\newblock In {\em ICLR}.

\bibitem[\protect\citeauthoryear{Elsken, Metzen, and Hutter}{2019b}]{survey}
Elsken, T.; Metzen, J.~H.; and Hutter, F.
\newblock 2019b.
\newblock Neural architecture search: {A} survey.
\newblock {\em J. Mach. Learn. Res.} 20:55:1--55:21.

\bibitem[\protect\citeauthoryear{Finn, Abbeel, and
  Levine}{2017}]{10.5555/3305381.3305498}
Finn, C.; Abbeel, P.; and Levine, S.
\newblock 2017.
\newblock Model-agnostic meta-learning for fast adaptation of deep networks.
\newblock In {\em ICML}.

\bibitem[\protect\citeauthoryear{Goldberg and
  Matari{\'{c}}}{2003}]{Goldberg2003}
Goldberg, D., and Matari{\'{c}}, M.~J.
\newblock 2003.
\newblock Maximizing reward in a non-stationary mobile robot environment.
\newblock {\em Autonomous Agents and Multi-Agent Systems} 6(3):287--316.

\bibitem[\protect\citeauthoryear{Hsu \bgroup et al\mbox.\egroup
  }{2018}]{hsu2018monas}
Hsu, C.-H.; Chang, S.-H.; Juan, D.-C.; Pan, J.-Y.; Chen, Y.-T.; Wei, W.; and
  Chang, S.-C.
\newblock 2018.
\newblock Monas: Multi-objective neural architecture search using reinforcement
  learning.
\newblock {\em arXiv preprint arXiv:1806.10332}.

\bibitem[\protect\citeauthoryear{Hu, Shen, and Sun}{2018}]{hu2018squeeze}
Hu, J.; Shen, L.; and Sun, G.
\newblock 2018.
\newblock Squeeze-and-excitation networks.
\newblock In {\em IEEE CVPR}.

\bibitem[\protect\citeauthoryear{{Huang} \bgroup et al\mbox.\egroup
  }{2017}]{8099726}
{Huang}, G.; {Liu}, Z.; v.~d. {Maaten}, L.; and {Weinberger}, K.~Q.
\newblock 2017.
\newblock Densely connected convolutional networks.
\newblock In {\em IEEE CVPR}.

\bibitem[\protect\citeauthoryear{Hutter, Hoos, and
  Leyton-Brown}{2011}]{10.1007/978-3-642-25566-3_40}
Hutter, F.; Hoos, H.~H.; and Leyton-Brown, K.
\newblock 2011.
\newblock Sequential model-based optimization for general algorithm
  configuration.
\newblock In {\em LION}.

\bibitem[\protect\citeauthoryear{Krizhevsky and
  Hinton}{2009}]{krizhevsky2009learning}
Krizhevsky, A., and Hinton, G.
\newblock 2009.
\newblock Learning multiple layers of features from tiny images.
\newblock Technical report, University of Toronto.

\bibitem[\protect\citeauthoryear{Laumanns \bgroup et al\mbox.\egroup
  }{2002}]{Laumanns:2002}
Laumanns, M.; Thiele, L.; Deb, K.; and Zitzler, E.
\newblock 2002.
\newblock Combining convergence and diversity in evolutionary multiobjective
  optimization.
\newblock {\em Evol. Comput.} 10(3):263--282.

\bibitem[\protect\citeauthoryear{Li \bgroup et al\mbox.\egroup
  }{2017}]{li2017metasgd}
Li, Z.; Zhou, F.; Chen, F.; and Li, H.
\newblock 2017.
\newblock Meta-sgd: Learning to learn quickly for few-shot learning.
\newblock {\em arXiv preprint arXiv:1707.09835}.

\bibitem[\protect\citeauthoryear{Liu \bgroup et al\mbox.\egroup
  }{2018}]{Liu_2018_ECCV}
Liu, C.; Zoph, B.; Neumann, M.; Shlens, J.; Hua, W.; Li, L.-J.; Fei-Fei, L.;
  Yuille, A.; Huang, J.; and Murphy, K.
\newblock 2018.
\newblock Progressive neural architecture search.
\newblock In {\em ECCV}.

\bibitem[\protect\citeauthoryear{Liu, Simonyan, and Yang}{2019}]{liu2018darts}
Liu, H.; Simonyan, K.; and Yang, Y.
\newblock 2019.
\newblock {DARTS}: Differentiable architecture search.
\newblock In {\em ICLR}.

\bibitem[\protect\citeauthoryear{Loshchilov and
  Hutter}{2017}]{loshchilov-ICLR17SGDR}
Loshchilov, I., and Hutter, F.
\newblock 2017.
\newblock Sgdr: Stochastic gradient descent with warm restarts.
\newblock In {\em ICLR}.

\bibitem[\protect\citeauthoryear{Lu \bgroup et al\mbox.\egroup
  }{2019}]{lu2018nsga}
Lu, Z.; Whalen, I.; Boddeti, V.; Dhebar, Y.; Deb, K.; Goodman, E.; and Banzhaf,
  W.
\newblock 2019.
\newblock Nsga-net: Neural architecture search using multi-objective genetic
  algorithm.
\newblock In {\em GECCO}.

\bibitem[\protect\citeauthoryear{Luo \bgroup et al\mbox.\egroup }{2018}]{NAO}
Luo, R.; Tian, F.; Qin, T.; Chen, E.; and Liu, T.-Y.
\newblock 2018.
\newblock Neural architecture optimization.
\newblock In {\em NeurIPS}.

\bibitem[\protect\citeauthoryear{Marler and Arora}{2004}]{Marler2004}
Marler, R., and Arora, J.
\newblock 2004.
\newblock Survey of multi-objective optimization methods for engineering.
\newblock {\em Structural and Multidisciplinary Optimization} 26(6):369--395.

\bibitem[\protect\citeauthoryear{{Parisi} \bgroup et al\mbox.\egroup
  }{2014}]{6889738}
{Parisi}, S.; {Pirotta}, M.; {Smacchia}, N.; {Bascetta}, L.; and {Restelli}, M.
\newblock 2014.
\newblock Policy gradient approaches for multi-objective sequential decision
  making.
\newblock In {\em Int. Joint Conf. on Neural Networks (IJCNN)}.

\bibitem[\protect\citeauthoryear{Parisi, Pirotta, and
  Restelli}{2016}]{Parisi:2016}
Parisi, S.; Pirotta, M.; and Restelli, M.
\newblock 2016.
\newblock Multi-objective reinforcement learning through continuous pareto
  manifold approximation.
\newblock {\em J. Artif. Int. Res.} 57(1):187--227.

\bibitem[\protect\citeauthoryear{Pham \bgroup et al\mbox.\egroup
  }{2018}]{pham2018efficient}
Pham, H.; Guan, M.~Y.; Zoph, B.; Le, Q.~V.; and Dean, J.
\newblock 2018.
\newblock Efficient neural architecture search via parameter sharing.
\newblock In {\em ICML}.

\bibitem[\protect\citeauthoryear{Real \bgroup et al\mbox.\egroup
  }{2019}]{Real_Aggarwal_Huang_Le_2019}
Real, E.; Aggarwal, A.; Huang, Y.; and Le, Q.~V.
\newblock 2019.
\newblock Regularized evolution for image classifier architecture search.
\newblock In {\em AAAI}.

\bibitem[\protect\citeauthoryear{Roijers \bgroup et al\mbox.\egroup
  }{2013}]{Roijers:2013}
Roijers, D.~M.; Vamplew, P.; Whiteson, S.; and Dazeley, R.
\newblock 2013.
\newblock A survey of multi-objective sequential decision-making.
\newblock {\em J. Artif. Int. Res.} 48(1):67--113.

\bibitem[\protect\citeauthoryear{{Sandler} \bgroup et al\mbox.\egroup
  }{2018}]{8578572}
{Sandler}, M.; {Howard}, A.; {Zhu}, M.; {Zhmoginov}, A.; and {Chen}, L.
\newblock 2018.
\newblock Mobilenetv2: Inverted residuals and linear bottlenecks.
\newblock In {\em IEEE CVPR}.

\bibitem[\protect\citeauthoryear{Srinivas and
  Deb}{1994}]{srinivas1994muiltiobjective}
Srinivas, N., and Deb, K.
\newblock 1994.
\newblock Muiltiobjective optimization using nondominated sorting in genetic
  algorithms.
\newblock {\em Evolutionary computation} 2(3):221--248.

\bibitem[\protect\citeauthoryear{Sutton and
  Barto}{2018}]{sutton2011reinforcement}
Sutton, R.~S., and Barto, A.~G.
\newblock 2018.
\newblock {\em Reinforcement Learning: An Introduction}.
\newblock Cambridge, MA: A Bradford Book.

\bibitem[\protect\citeauthoryear{Tan \bgroup et al\mbox.\egroup
  }{2019}]{tan2018mnasnet}
Tan, M.; Chen, B.; Pang, R.; Vasudevan, V.; Sandler, M.; Howard, A.; and Le,
  Q.~V.
\newblock 2019.
\newblock Mnasnet: Platform-aware neural architecture search for mobile.
\newblock In {\em IEEE CVPR}.

\bibitem[\protect\citeauthoryear{Wei \bgroup et al\mbox.\egroup
  }{2016}]{Wei:2016:NM:3045390.3045451}
Wei, T.; Wang, C.; Rui, Y.; and Chen, C.~W.
\newblock 2016.
\newblock Network morphism.
\newblock In {\em Proc. Int. Conf. Machine Learning}.

\bibitem[\protect\citeauthoryear{Williams}{1992}]{Williams:1992:SSG:139611.139614}
Williams, R.~J.
\newblock 1992.
\newblock Simple statistical gradient-following algorithms for connectionist
  reinforcement learning.
\newblock {\em Mach. Learn.} 8(3-4):229--256.

\bibitem[\protect\citeauthoryear{Wu \bgroup et al\mbox.\egroup
  }{2019}]{wu2018fbnet}
Wu, B.; Dai, X.; Zhang, P.; Wang, Y.; Sun, F.; Wu, Y.; Tian, Y.; Vajda, P.;
  Jia, Y.; and Keutzer, K.
\newblock 2019.
\newblock Fbnet: Hardware-aware efficient convnet design via differentiable
  neural architecture search.
\newblock In {\em IEEE CVPR}.

\bibitem[\protect\citeauthoryear{Xie \bgroup et al\mbox.\egroup
  }{2019}]{xie2018snas}
Xie, S.; Zheng, H.; Liu, C.; and Lin, L.
\newblock 2019.
\newblock {SNAS}: stochastic neural architecture search.
\newblock In {\em ICLR}.

\bibitem[\protect\citeauthoryear{Yang, Esperança, and
  Carlucci}{2020}]{yang2019nas}
Yang, A.; Esperança, P.~M.; and Carlucci, F.~M.
\newblock 2020.
\newblock Nas evaluation is frustratingly hard.
\newblock In {\em ICLR}.

\bibitem[\protect\citeauthoryear{{Yu} and {Mannor}}{2009}]{5137416}
{Yu}, J.~Y., and {Mannor}, S.
\newblock 2009.
\newblock Online learning in markov decision processes with arbitrarily
  changing rewards and transitions.
\newblock In {\em Int. Conf. Game Theory for Networks}.

\bibitem[\protect\citeauthoryear{Zhang, Ren, and
  Urtasun}{2019}]{zhang2018graph}
Zhang, C.; Ren, M.; and Urtasun, R.
\newblock 2019.
\newblock Graph hypernetworks for neural architecture search.
\newblock In {\em ICLR}.

\bibitem[\protect\citeauthoryear{{Zitzler} \bgroup et al\mbox.\egroup
  }{2003}]{1197687}
{Zitzler}, E.; {Thiele}, L.; {Laumanns}, M.; {Fonseca}, C.~M.; and {da
  Fonseca}, V.~G.
\newblock 2003.
\newblock Performance assessment of multiobjective optimizers: an analysis and
  review.
\newblock {\em IEEE Trans. Evol. Comp.} 7(2):117--132.

\bibitem[\protect\citeauthoryear{Zoph and Le}{2017}]{zoph2017neural}
Zoph, B., and Le, Q.~V.
\newblock 2017.
\newblock Neural architecture search with reinforcement learning.
\newblock In {\em ICLR}.

\bibitem[\protect\citeauthoryear{Zoph \bgroup et al\mbox.\egroup
  }{2018}]{zoph2018learning}
Zoph, B.; Vasudevan, V.; Shlens, J.; and Le, Q.~V.
\newblock 2018.
\newblock Learning transferable architectures for scalable image recognition.
\newblock In {\em IEEE CVPR}.

\end{thebibliography}
\bibliographystyle{aaai}

\section{Appendix}

\subsection{Scatter Plots of Sampled Architectures}
\label{appendix:scatter_plots}
\begin{figure}[ht!]
	\centering
	\begin{subfigure}[t]{0.23\textwidth}
		\centering
		\includegraphics[width=\textwidth]{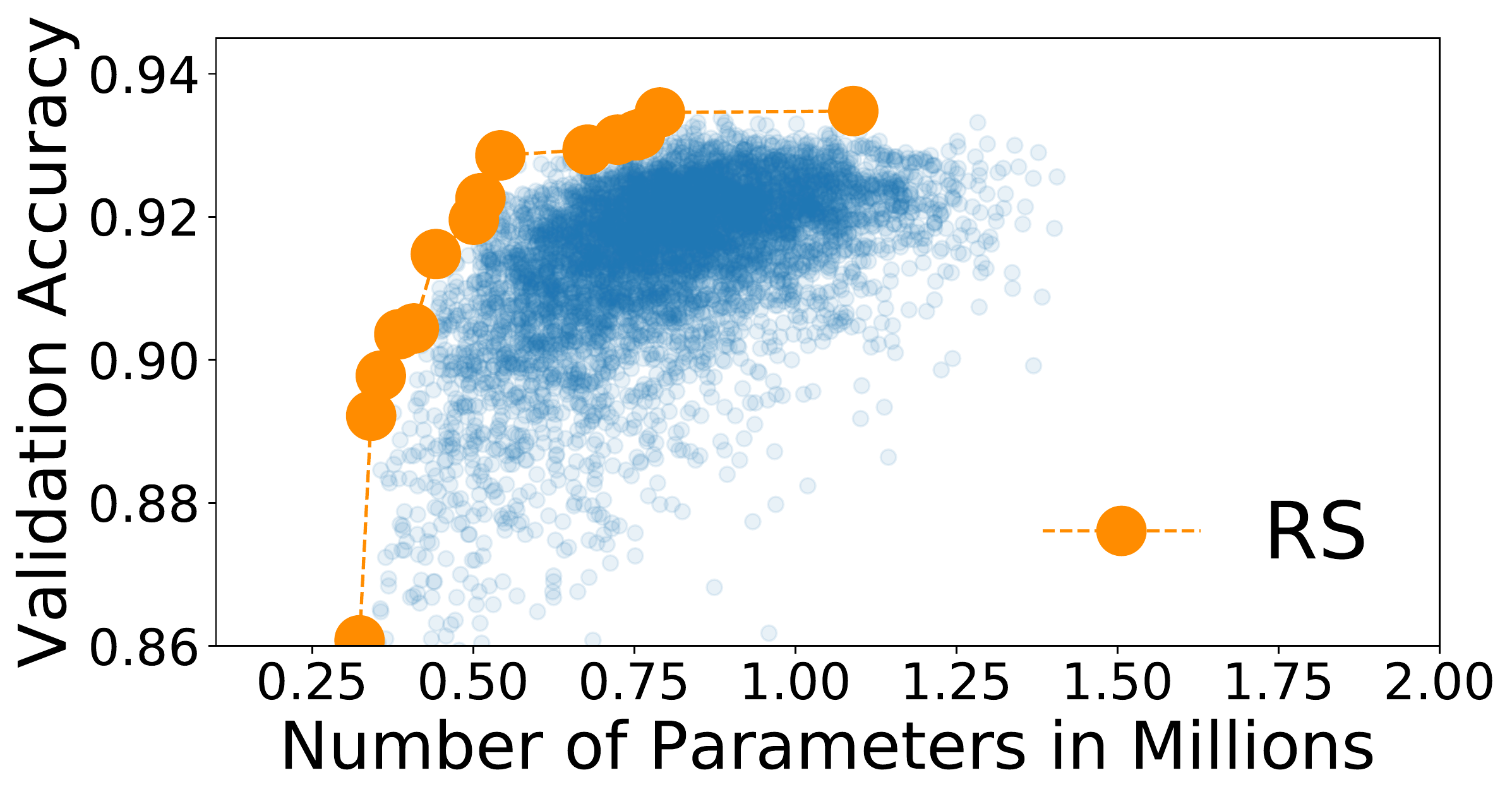}
		\caption{}
	\end{subfigure}
	\begin{subfigure}[t]{0.23\textwidth}
		\centering
		\includegraphics[width=\textwidth]{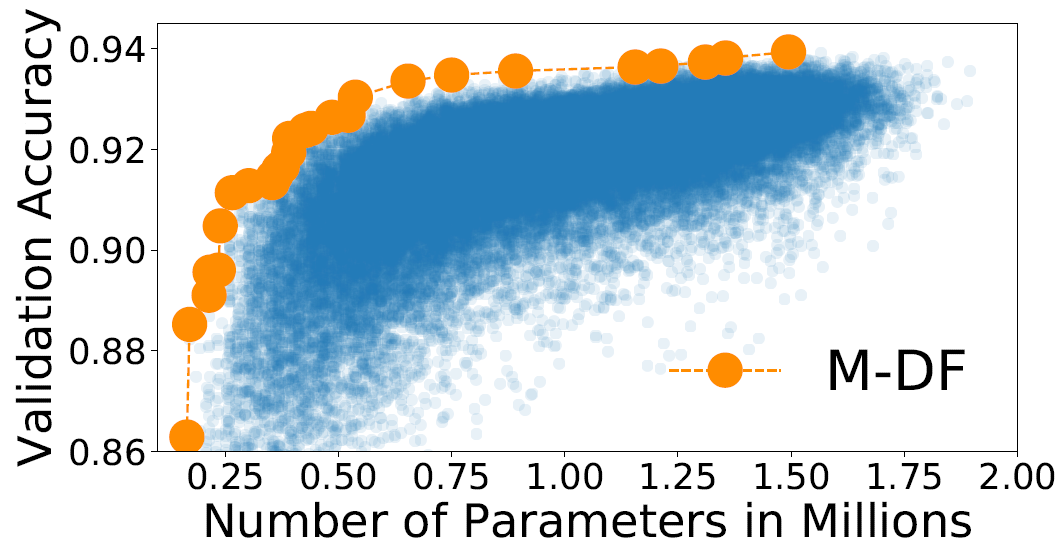}
		\caption{}
	\end{subfigure}
    \begin{subfigure}[t]{0.23\textwidth}
    	\centering
    	\includegraphics[width=\textwidth]{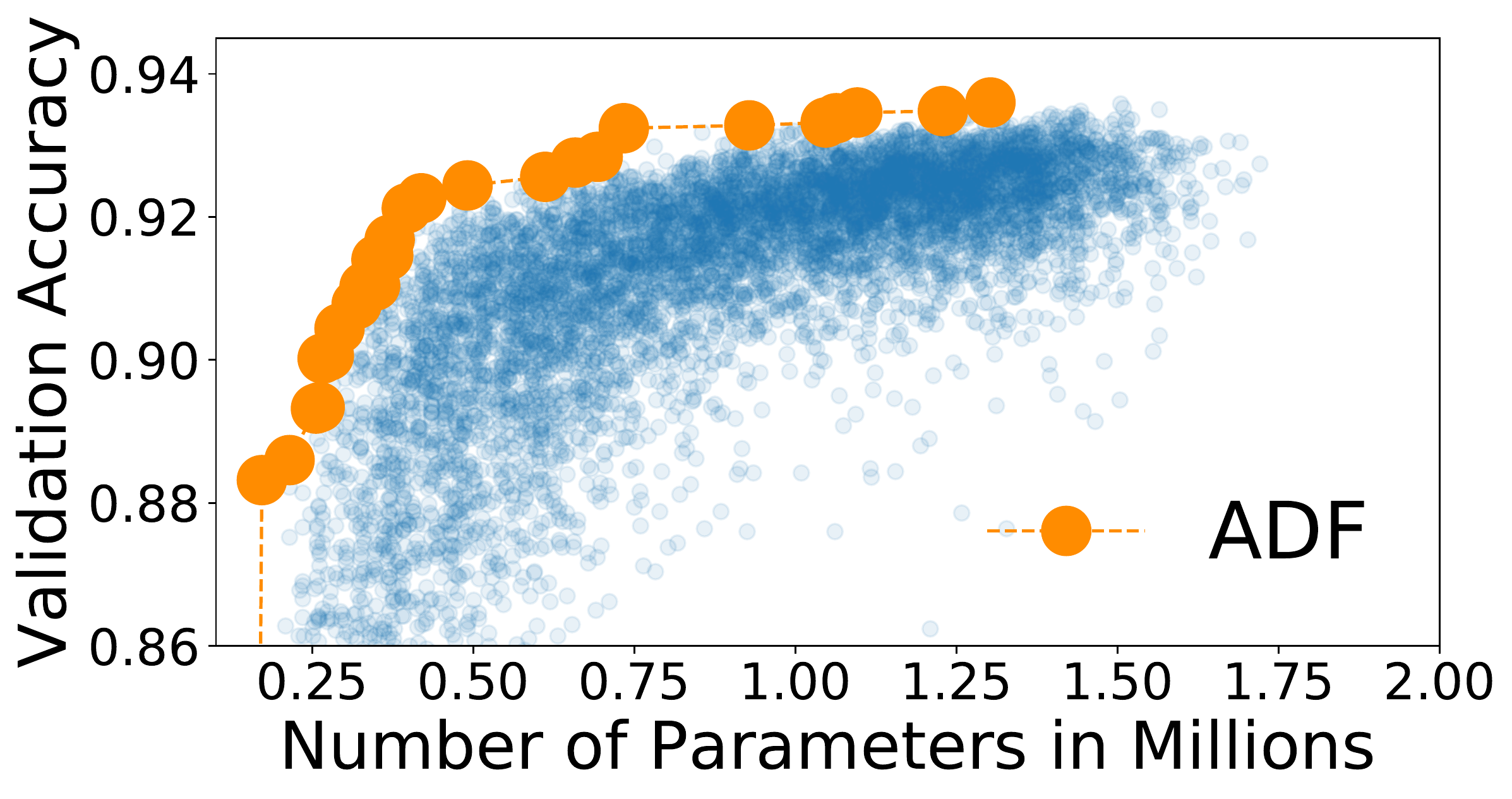}
    	\caption{}
    \end{subfigure}
    \begin{subfigure}[t]{0.23\textwidth}
    	\centering
    	\includegraphics[width=\textwidth]{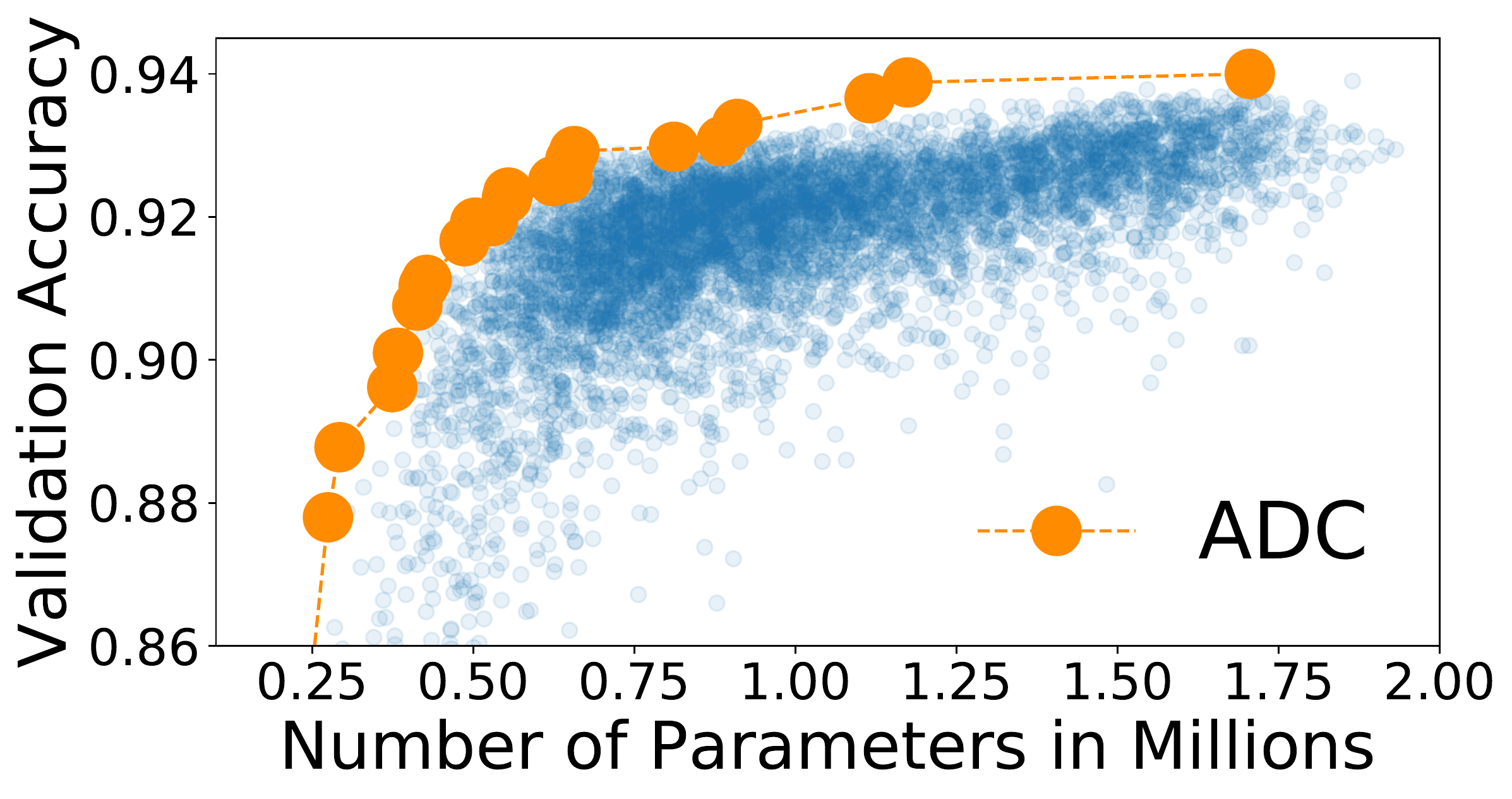}
    	\caption{}
    \end{subfigure}
    \begin{subfigure}[t]{0.23\textwidth}
    	\centering
    	\includegraphics[width=\textwidth]{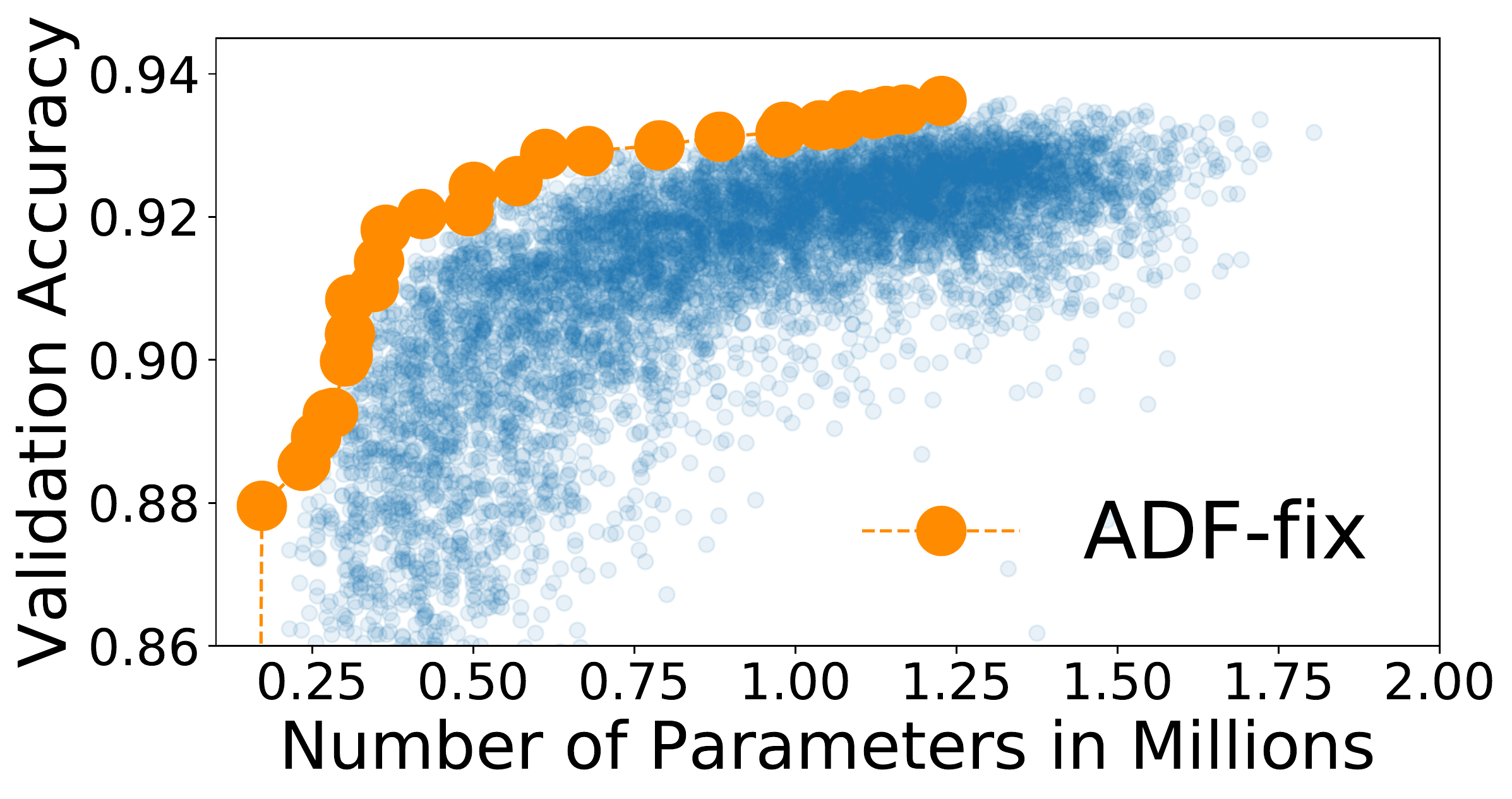}
    	\caption{}
    \end{subfigure}
    \begin{subfigure}[t]{0.23\textwidth}
    	\centering
    	\includegraphics[width=\textwidth]{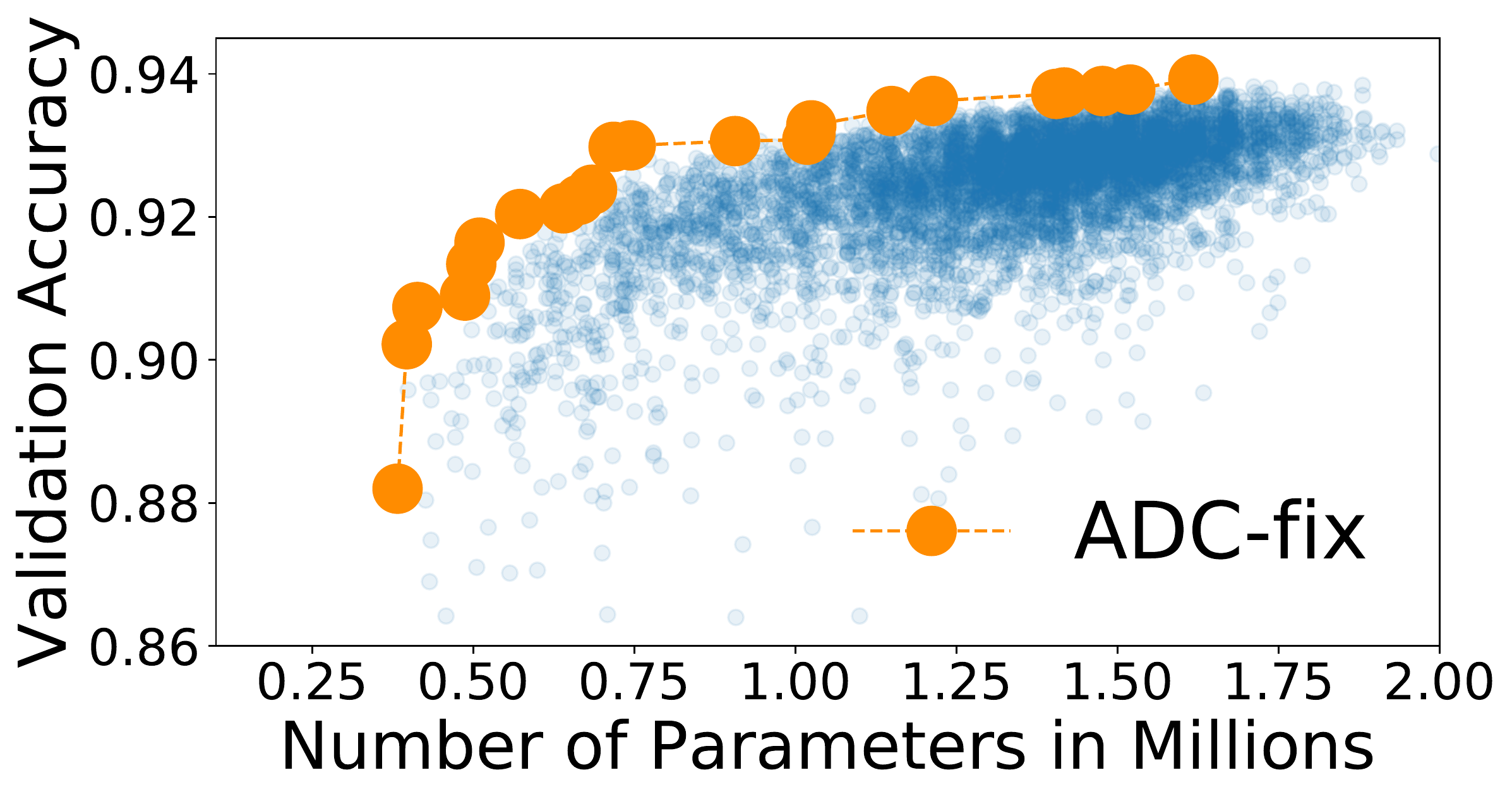}
    	\caption{}
    \end{subfigure}
	\caption{Scatter plots for architectures sampled by (a) RS, (b) M-DF, (c) ADF, (d) ADC, (e) ADF with a fixed temperature $T=5$, and (f) ADC with a fixed temperature $T=5$.}
	\label{fig:sample_region}
\end{figure}

We plot in Fig. \ref{fig:sample_region} all architectures sampled by different methods for comparison purposes. We observe that M-DF, ADF, ADF-fix and ADC sample the search space more uniformly than other methods. RS has poor sampling behavior since it oversamples a small part of the subspace. 
Finally, we confirm that ADF and ADF-fix behave quite similarly, while there is a big gap in the performance of ADC and ADC-fix. Concretely, the variant with improved exploration may be more likely to avoid getting stuck in a specific subspace, and thus expected to explore more parts of the search space.

\subsection{Discovered Cells}
\label{appendix:discovered_cells}
\begin{figure}[ht!]
	\centering
	\begin{subfigure}[h]{0.44\textwidth}
		\centering
		\includegraphics[width=\textwidth,keepaspectratio]{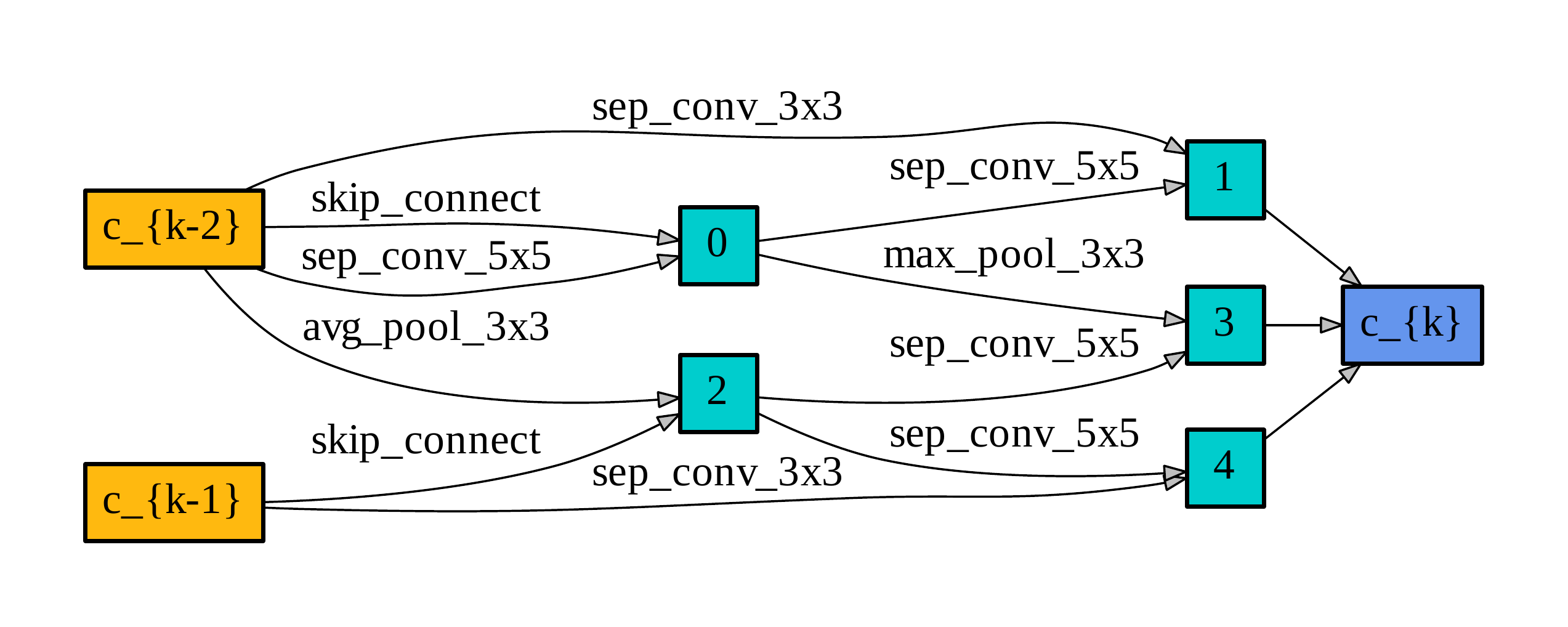}
		\caption{Normal cell of ADF-L10.}
	\end{subfigure}
	\begin{subfigure}[h]{0.44\textwidth}
		\centering
		\includegraphics[width=\textwidth,keepaspectratio]{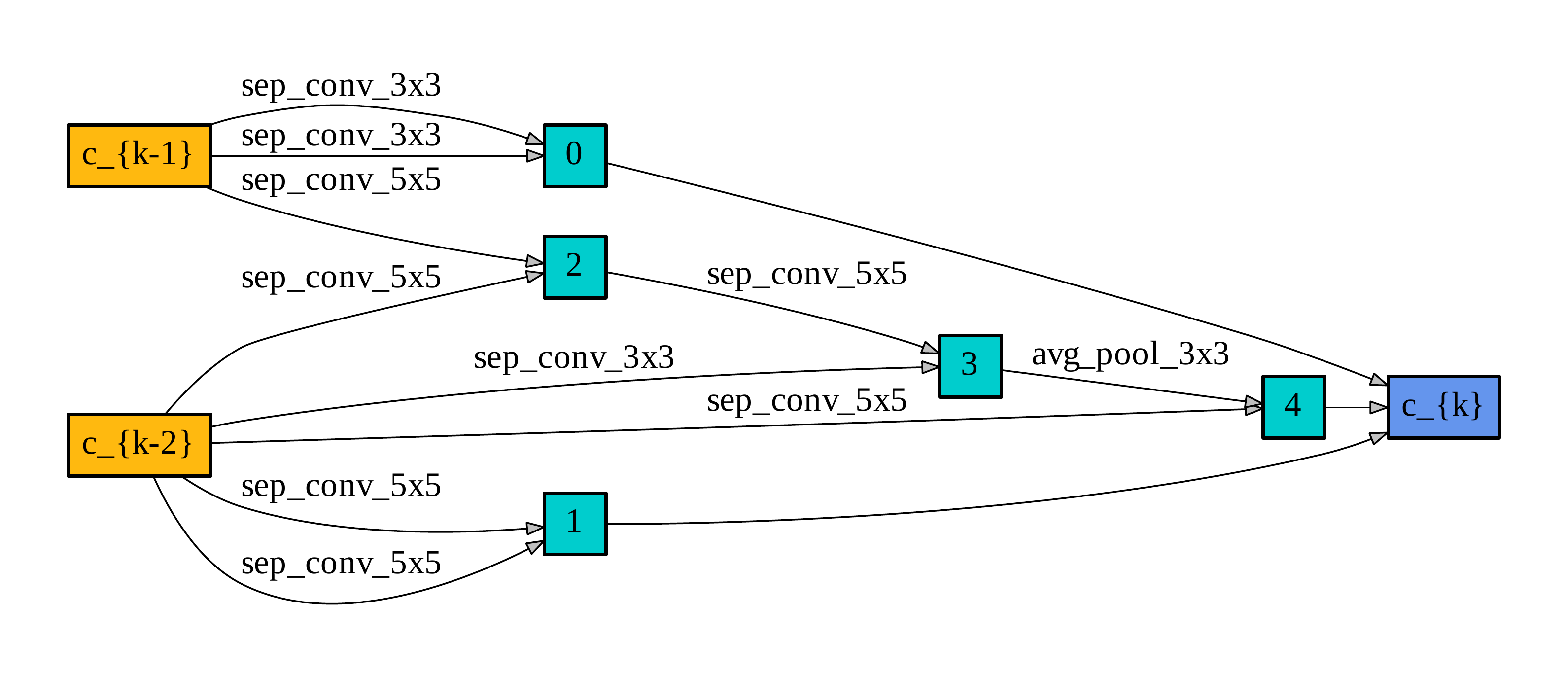}
		\caption{Reduction cell of ADF-L10.}
	\end{subfigure}
	\caption{Cells for ADF-L10.}
	\label{fig:ADF_L10}
	
\end{figure}

\begin{figure}[ht!]
	\centering
	\begin{subfigure}[h]{0.44\textwidth}
		\centering
		\includegraphics[width=\textwidth,keepaspectratio]{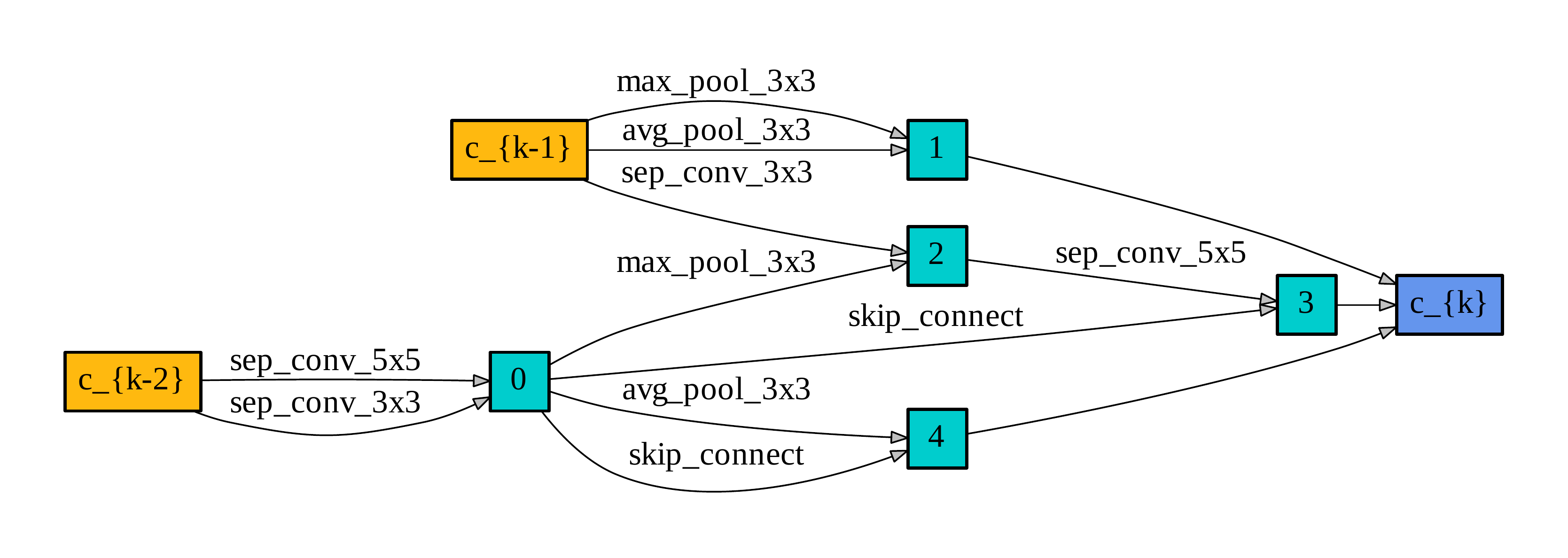}
		\caption{Normal cell of ADF-M10.}
	\end{subfigure}
	\begin{subfigure}[h]{0.44\textwidth}
		\centering
		\includegraphics[width=\textwidth,keepaspectratio]{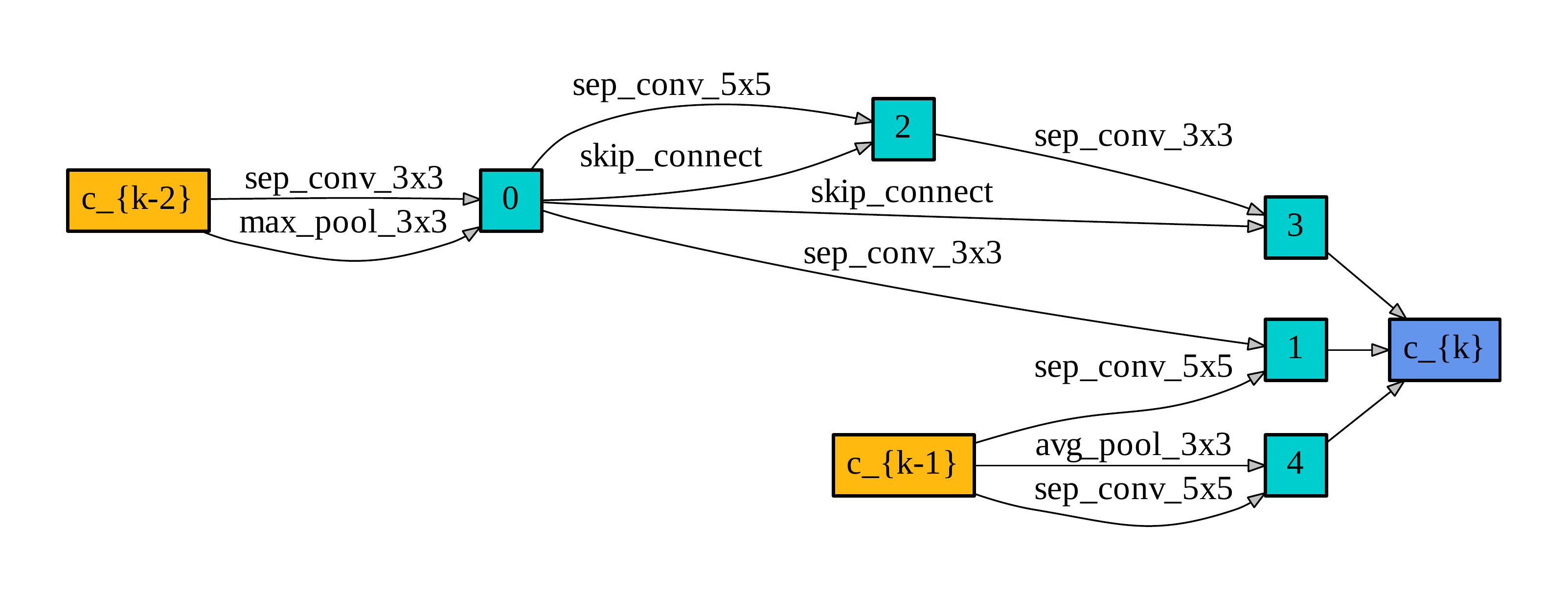}
		\caption{Reduction cell of ADF-M10.}
	\end{subfigure}
	\caption{Cells for ADF-M10.}
	\label{fig:ADF_M10}
\end{figure}

\begin{figure}[ht!]
	\centering
	\begin{subfigure}[h]{0.44\textwidth}
		\centering
		\includegraphics[width=\textwidth, keepaspectratio]{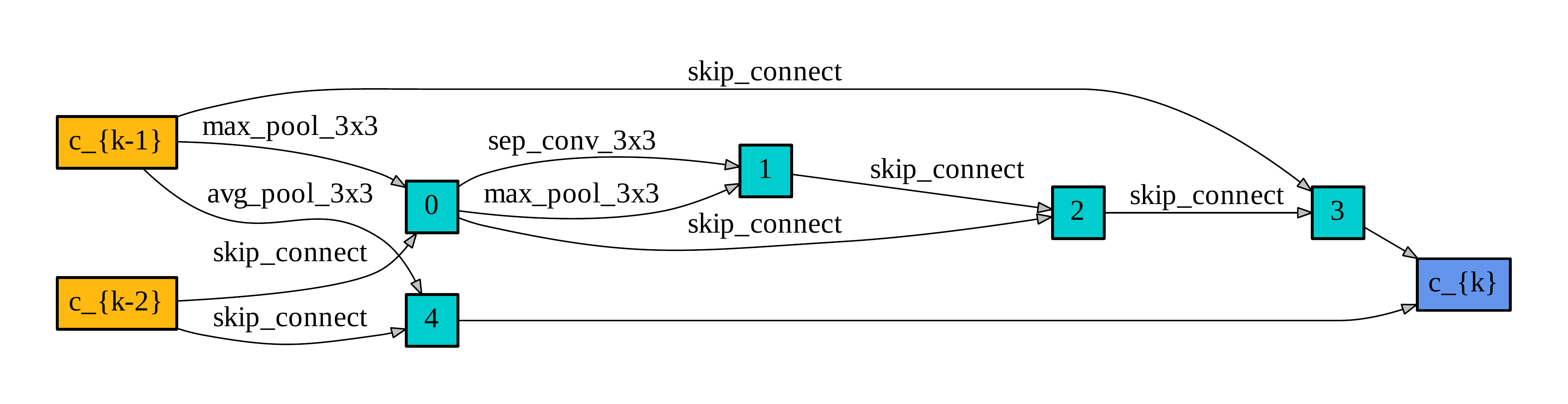}
		\caption{Normal cell of ADF-S10.}
	\end{subfigure}
	\begin{subfigure}[h]{0.44\textwidth}
		\centering
		\includegraphics[width=\textwidth,keepaspectratio]{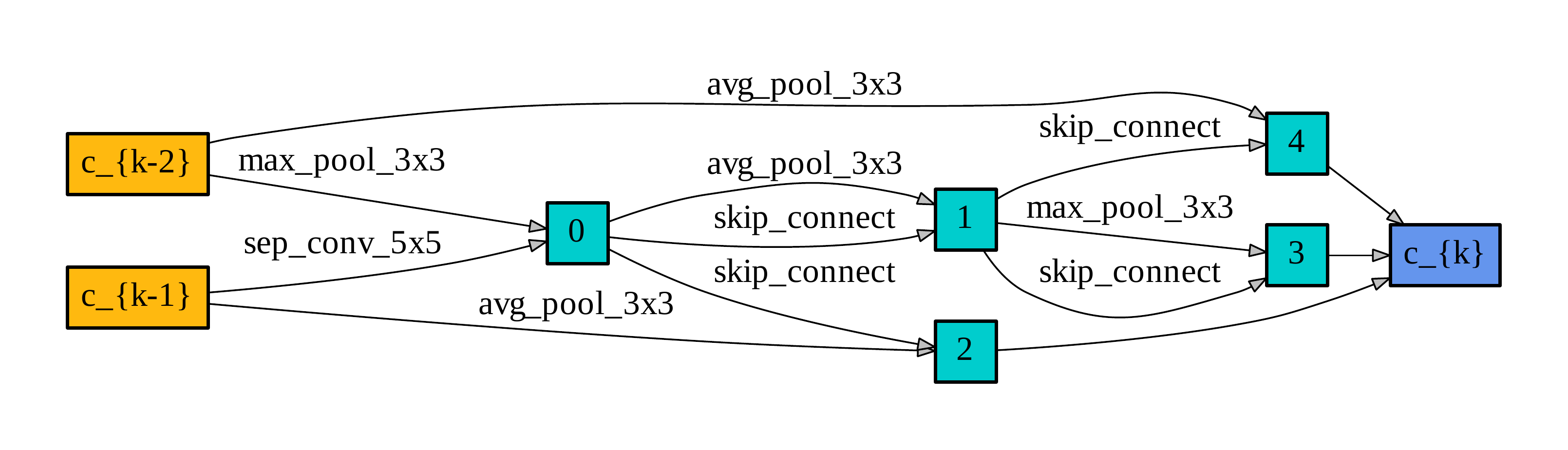}
		\caption{Reduction cell of ADF-S10.}
	\end{subfigure}
	\caption{Cells for ADF-S10.}
	\label{fig:ADF_S10}
\end{figure}

\begin{figure}[ht!]
	\centering
	\begin{subfigure}[t]{0.23\textwidth}
		\centering
		\includegraphics[width=\textwidth]{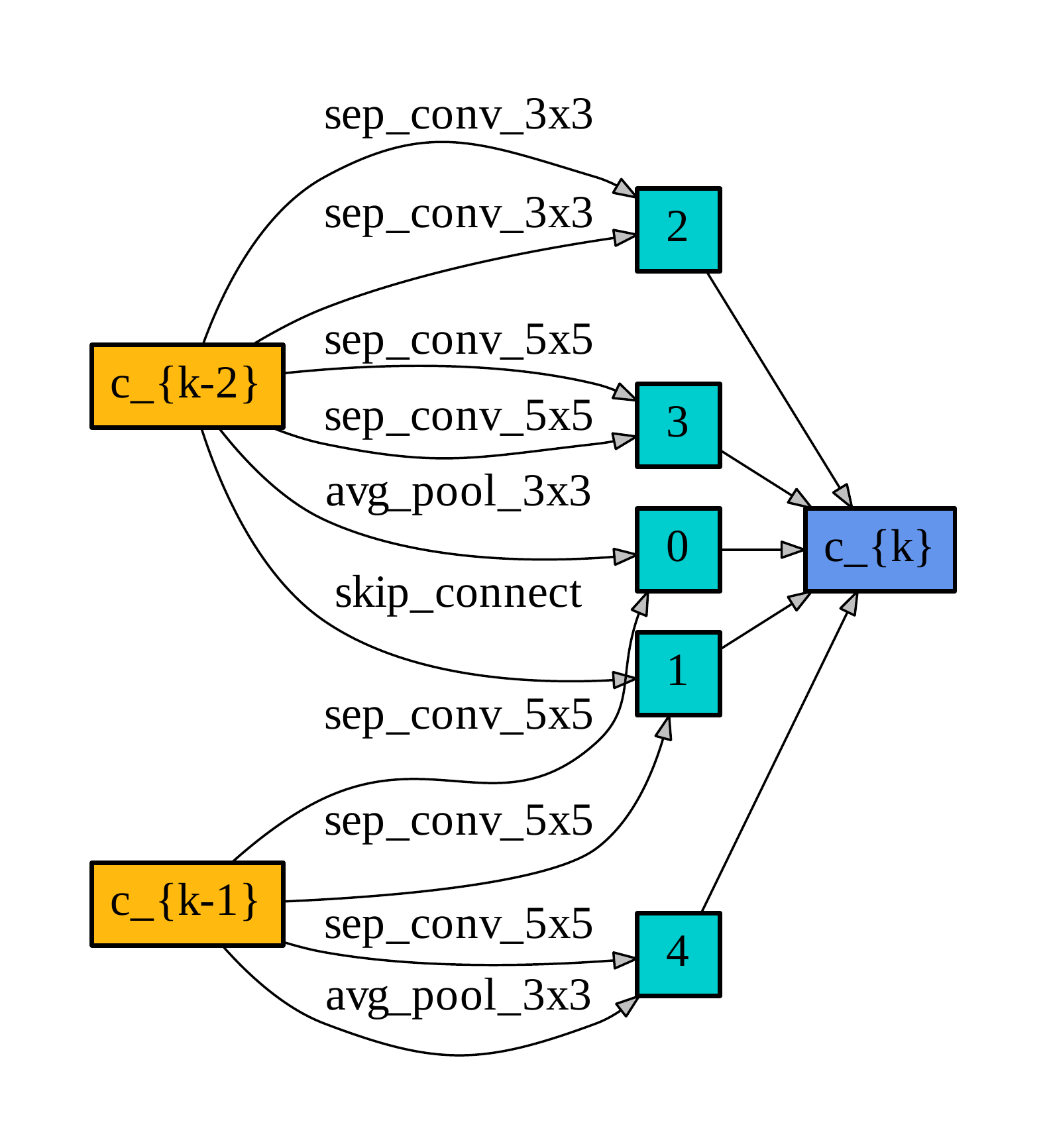}
		\caption{Normal cell of ADC-L10.}
	\end{subfigure}
	\begin{subfigure}[t]{0.23\textwidth}
		\centering
		\includegraphics[width=\textwidth]{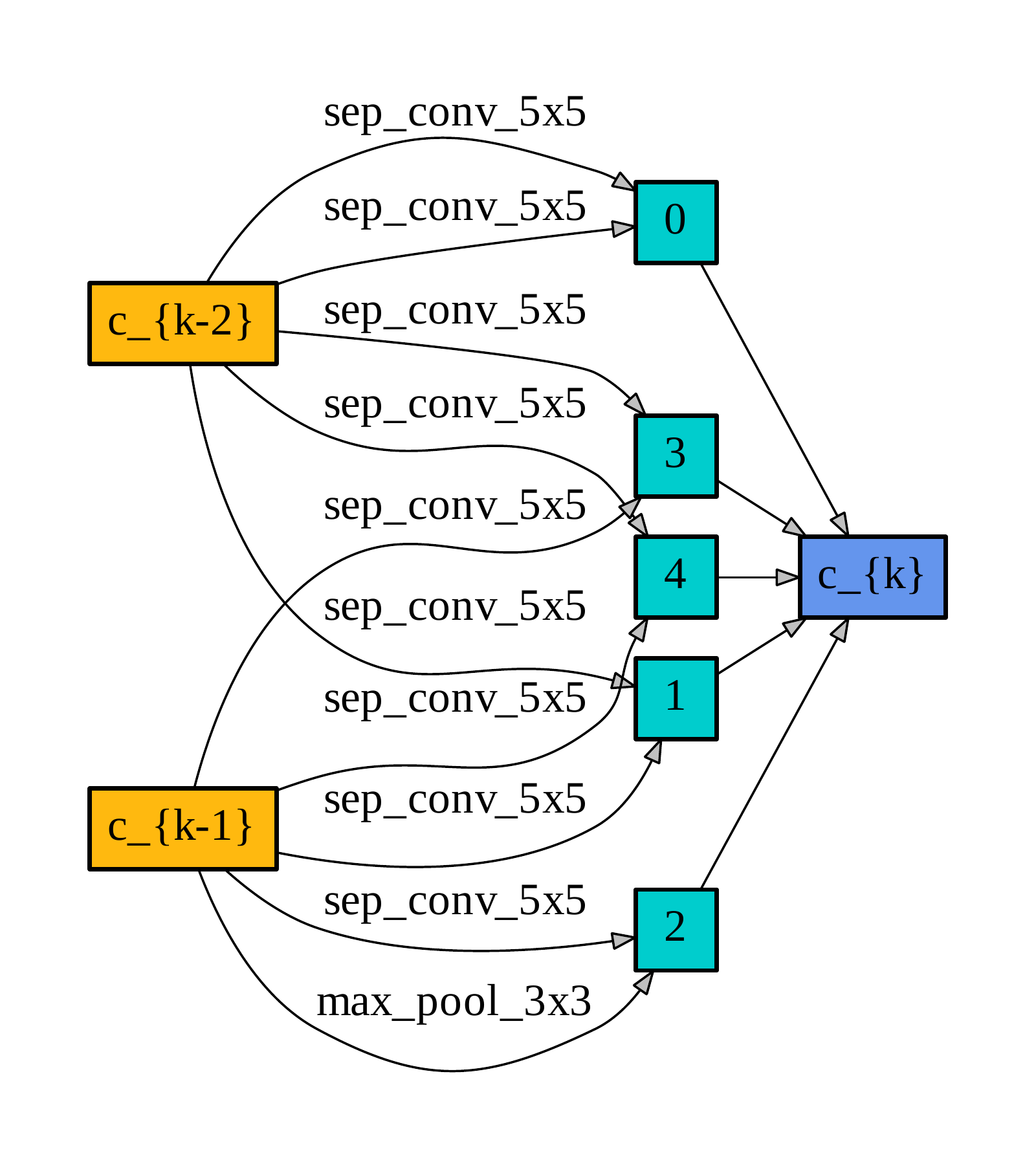}
		\caption{Reduction cell of ADC-L10.}
	\end{subfigure}
	\caption{Cells for ADC-L10.}
	\label{fig:ADC_L10}
\end{figure}

\begin{figure}[ht!]
	\centering
	\begin{subfigure}[t]{0.44\textwidth}
		\centering
		\includegraphics[width=\textwidth,keepaspectratio]{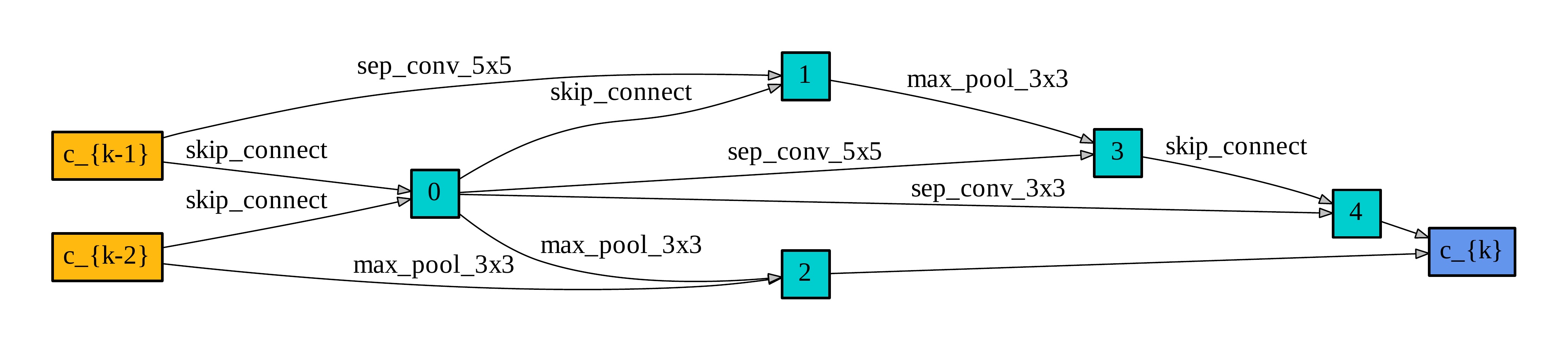}
		\caption{Normal cell of ADC-M10.}
	\end{subfigure}
	\begin{subfigure}[t]{0.44\textwidth}
		\centering
		\includegraphics[width=\textwidth,keepaspectratio]{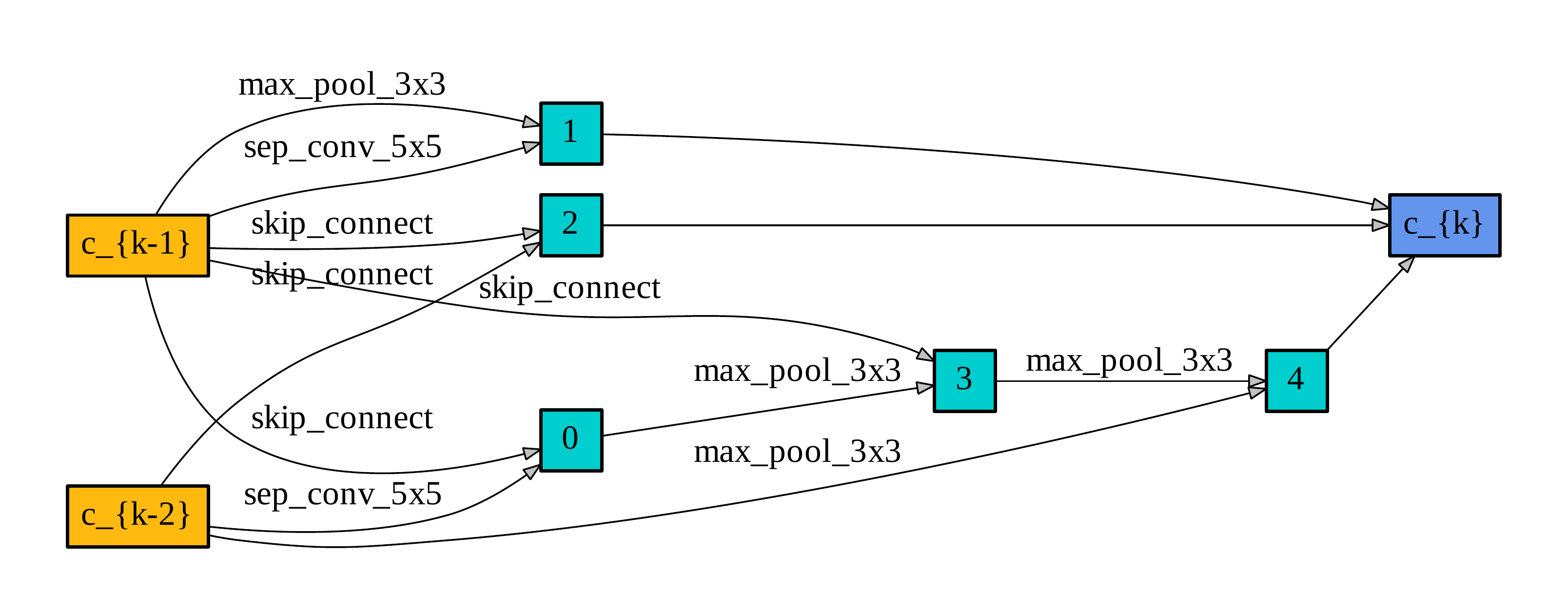}
		\caption{Reduction cell of ADC-M10.}
	\end{subfigure}
	\caption{Cells for ADC-M10.}
	\label{fig:ADC_M10}
\end{figure}

\begin{figure}[ht!]
	\centering
	\begin{subfigure}[t]{0.44\textwidth}
		\centering
		\includegraphics[width=\textwidth,keepaspectratio]{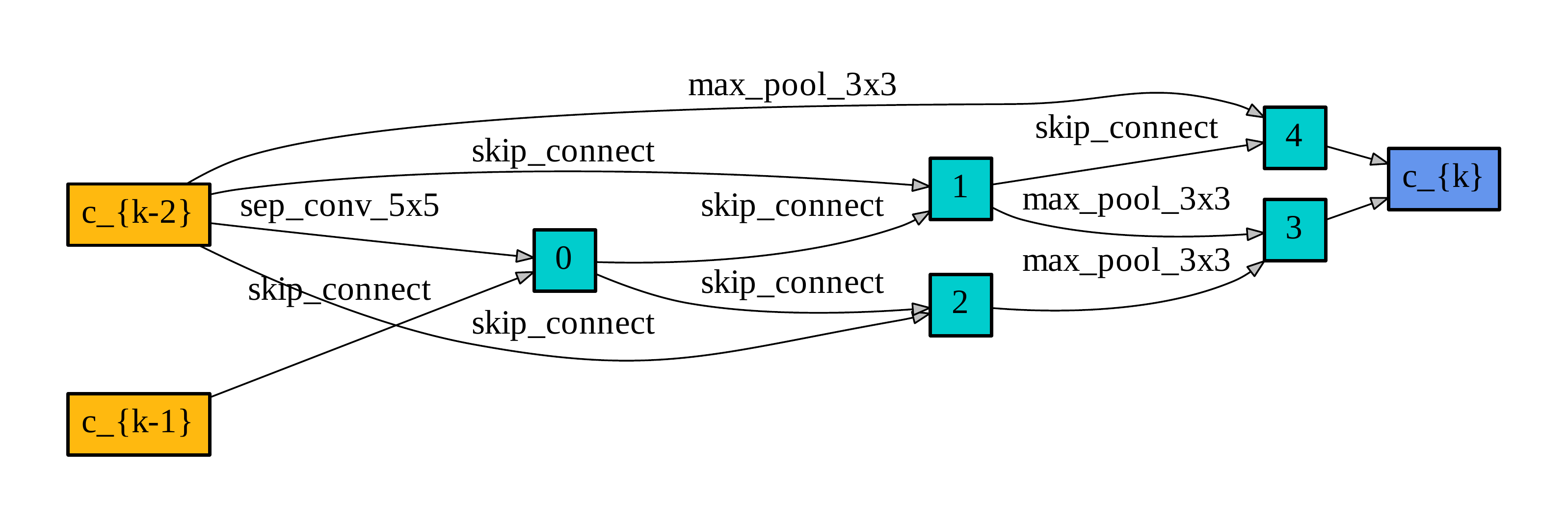}
		\caption{Normal cell of ADC-S10.}
	\end{subfigure}
	\begin{subfigure}[t]{0.44\textwidth}
		\centering
		\includegraphics[width=\textwidth]{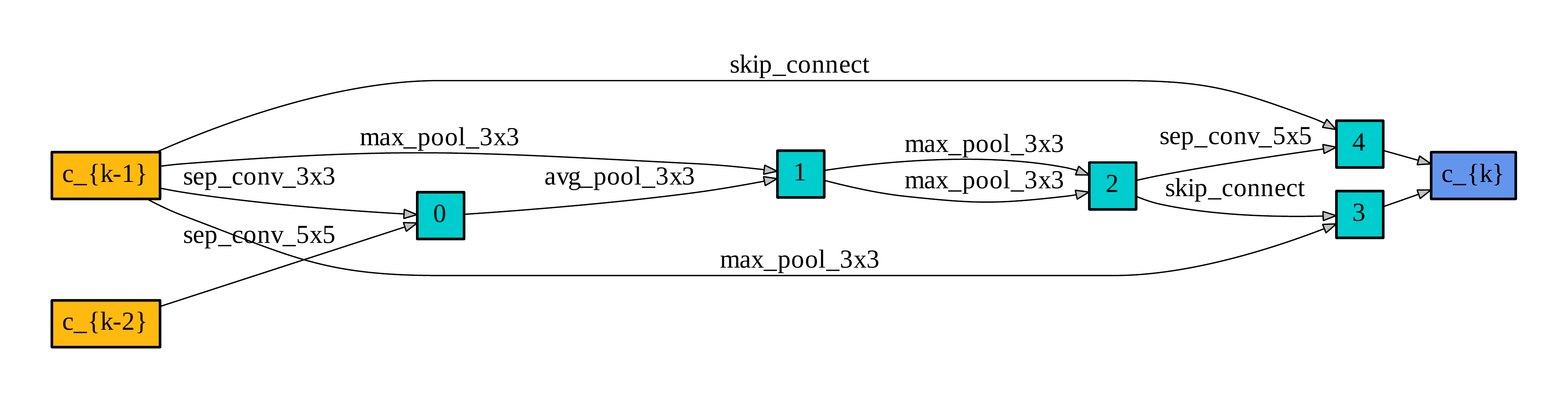}
		\caption{Reduction cell of ADC-S10.}
	\end{subfigure}
	\caption{Cells for ADC-S10.}
	\label{fig:ADC_S10}
\end{figure}

Various NPG-NAS cells for CIFAR-10 with varying numbers of parameters (large, medium, or small) are visualized in Figs.~\ref{fig:ADF_L10}-\ref{fig:ADC_S10}. 
We observe that the cells produced by our search process have very diverse structures.
As expected, cells with more parameters tend to have more convolution operators, while as the number of parameters decreases, cells get more skip connections and pooling operators.
Note that cells associated with fewer parameters tend to be deeper (this is more pronounced for normal cells).


\end{document}